\documentclass[sigconf]{acmart} 

\acmConference[SenSys '26]{The International Conference on Embedded Artificial Intelligence and Sensing Systems}{November 2026}{TBD}
\acmYear{2026}\copyrightyear{2026}
\acmISBN{TBD}\acmDOI{TBD}
\usepackage{graphicx}
\usepackage{booktabs}
\usepackage{multirow}
\usepackage{array}
\usepackage{tabularx}
\usepackage{float}
\usepackage{subcaption}
\usepackage{amsmath}
\usepackage{xcolor}
\usepackage{tikz}
\usepackage{pgfplots}
\usepackage{balance}   
\usepackage{enumitem}
\usepackage[autostyle,english=american]{csquotes} 
\usepackage{microtype}
\usepackage{subcaption}
\usepackage{forest}
\usepackage{siunitx}
\usepackage[capitalise,nameinlink,noabbrev]{cleveref}

\crefname{figure}{Fig.}{Figs.}
\Crefname{figure}{Fig.}{Figs.}
\crefname{table}{Table}{Tables}
\Crefname{table}{Table}{Tables}

\pgfplotsset{compat=1.18}

\definecolor{ACTbase}{RGB}{41,128,185}    
\definecolor{OVLAbase}{RGB}{142,68,173}   
\definecolor{RDTbase}{RGB}{26,188,156}    
\definecolor{PI0base}{RGB}{243,156,18}    

\pgfplotsset{
  resultsplotrow/.style={
    ybar,
    bar width=7pt,
    width=\columnwidth,
    height=5cm,
    ymin=0, ymax=100,
    ylabel={Success Rate (\%)},
    symbolic x coords={Clean Dish, Put into Pot, Unzip Bag, Folding Shorts (200)},
    xtick=data,
    xticklabel style={align=center, text width=1.65cm},
    enlarge x limits=0.20,   
    clip=false,              
    ymajorgrids,
    grid style=dashed,
    nodes near coords,
    nodes near coords style={font=\scriptsize},
    legend style={draw=none, fill=none, font=\footnotesize, at={(0.98,0.98)}, anchor=north east},
    tick label style={font=\small},
    label style={font=\small},
  }
}

\forestset{
  vla-tax-lr/.style={
    for tree={
      draw, rounded corners, align=left, font=\footnotesize,
      inner sep=2.5pt, l sep=16pt, s sep=12pt,
      grow'=east,               
      parent anchor=east, child anchor=west, anchor=west,
      edge path={               
        \noexpand\path[draw] (!u.parent anchor) -- +(6pt,0) |- (.child anchor)
        \forestoption{edge label};
      }
    },
    where level=0{fill=gray!15, font=\footnotesize\bfseries, inner sep=4pt}{},
    where level=1{font=\footnotesize\bfseries}{}
  }
}

\title{Experiences from Benchmarking Vision–Language–Action Models for Robotic Manipulation}

\author{Yihao Zhang}
\affiliation{%
  \institution{Macquarie University}
  \city{Sydney}
  \state{NSW}
  \country{Australia}
}
\email{yihao.zhang3@hdr.mq.edu.au}

\author{Yuankai Qi}
\affiliation{%
  \institution{Macquarie University}
  \city{Sydney}
  \state{NSW}
  \country{Australia}
}
\email{yuankai.qi@mq.edu.au}

\author{Xi Zheng}
\affiliation{%
  \institution{Macquarie University}
  \city{Sydney}
  \state{NSW}
  \country{Australia}
}
\email{james.zheng@mq.edu.au}



\begin{document}

\begin{abstract} 
Foundation models applied in robotics, particularly \textbf{Vision--Language--Action (VLA)} models, hold great promise for achieving general-purpose manipulation. Yet, systematic real-world evaluations and cross-model comparisons remain scarce. This paper reports our \textbf{empirical experiences} from benchmarking four representative VLAs---\textbf{ACT}, \textbf{OpenVLA--OFT}, \textbf{RDT-1B}, and \boldmath{$\pi_0$}---across four manipulation tasks conducted in both simulation and on the \textbf{ALOHA Mobile} platform. We establish a \textbf{standardized evaluation framework} that measures performance along three key dimensions: (1) \textit{accuracy and efficiency} (success rate and time-to-success), (2) \textit{adaptability} across in-distribution, spatial out-of-distribution, and instance-plus-spatial out-of-distribution settings, and (3) \textit{language instruction-following accuracy}. Through this process, we observe that \boldmath{$\pi_0$} demonstrates superior adaptability in out-of-distribution scenarios, while \textbf{ACT} provides the highest stability in-distribution. Further analysis highlights differences in computational demands, data-scaling behavior, and recurring failure modes such as near-miss grasps, premature releases, and long-horizon state drift. These findings reveal practical trade-offs among VLA model architectures in balancing precision, generalization, and deployment cost, offering actionable insights for selecting and deploying VLAs in real-world robotic manipulation tasks.
\end{abstract}

\maketitle

\section{Introduction}
Robotic manipulation aims to enable robots to grasp, move, and assemble objects for automation across manufacturing, healthcare, and service domains \cite{li2024real,firoozi2023foundation}. While traditional task-specific controllers achieve precise performance in structured settings, they lack adaptability to new or unstructured tasks \cite{siciliano2016handbook,mason2001mechanics}. Imitation learning (IL) has improved generalization by enabling robots to learn manipulation skills from human demonstrations \cite{osa2018algorithmic}, with methods such as behavior cloning (BC) and inverse reinforcement learning (IRL). Despite progress, IL-based systems still struggle to generalize beyond training distributions—particularly under object, spatial, or environmental variations—highlighting the need for foundation models that integrate vision, language, and action for more scalable and robust robotic learning.

\noindent\textbf{Motivation and Prior Landscape.}
Scaling capacity in Vision--Language--Action (VLA) models can improve generalization, but it amplifies data, compute, and real-time control trade-offs \cite{firoozi2023foundation}. High-capacity VLAs need large, diverse datasets across tasks, language commands, and environments; for example, RT-1 gathered 130{,}000 demonstrations over 17 months on 13 robots \cite{singh2023scaling}. Synthetic data and simulation reduce cost \cite{singh2023scaling}, yet ensuring realism and diversity is difficult. Latency from large vision encoders and (for tokenized policies) autoregressive decoding further pressures 10--50\,Hz control budgets \cite{brohan2023rt2,openvla,openvla_oft}; optimizations such as token caching, low-rank adaptation, and quantization help but do not fully resolve on-board constraints \cite{vla_cache,dao2023flashattention2}. Despite rapid progress, SOTA evaluations often target single-arm, simulation-only setups and report mixed robustness under distribution shifts (20\%--70\% drops on new objects, layouts, or viewpoints) \cite{wang2024towards,liu2025evavla}, leaving open questions about real-world reliability, instruction following, and spatial generalization.

\noindent\textbf{Our Benchmark and Design Choices.}
We present an experience-driven benchmark that directly compares representative models under identical settings on complex dual-arm, household-inspired tasks requiring coordinated two-handed control, tool use, deformable-object handling, and language grounding. We evaluate a specialist imitation-learning policy (ACT \cite{zhao2023act}) against three generalist VLAs (OpenVLA--OFT \cite{openvla_oft}, RDT-1B \cite{rdt1b}, and $\pi_0$ \cite{pi0}), chosen to span diverse architectures, training paradigms, and community relevance. Experiments run both in simulation (for scale, ablations, and safety) and on the ALOHA Mobile platform (to capture sensing, latency, and actuation constraints), ensuring that results reflect practical deployment realities rather than simulation artifacts.

\noindent\textbf{Metrics, Settings, and Insights.}
Our standardized evaluation emphasizes (i) \emph{accuracy and efficiency}: success rate and time-to-success; (ii) \emph{adaptability}: performance under \emph{in-distribution (ID)} conditions, \emph{spatial OOD} (same instances with new placements, poses, or layouts), and \emph{instance{+}spatial OOD} (new object instances \emph{and} new spatial configurations); and (iii) \emph{language instruction-following accuracy}. This design isolates precision vs.\ generalization vs.\ deployment cost. Empirically, $\pi_0$ shows strong adaptability under OOD changes, while ACT is most stable in-distribution; we also document recurrent failure modes (near-miss grasps, premature release, long-horizon state drift) and compute/latency trade-offs across models. These insights offer actionable guidance on when to prefer specialist vs.\ generalist policies and how evaluation in both real and simulated settings reveals complementary facets of VLA performance.

\noindent\textbf{Contributions.} This paper provides the first systematic empirical evaluation of generalist Vision–Language–Action (VLA) models on real-world bimanual manipulation. Our main contributions are:

\begin{itemize}
    \item \textbf{Unified Benchmark:} 
    We introduce a dual-arm manipulation benchmark with well-defined in-distribution and out-of-distribution evaluation settings, and release a standardized, reproducible evaluation pipeline with unified metrics for fair cross-model comparison.
    
    \item \textbf{Failure Taxonomy and Diagnostic Analysis:} 
    We develop a structured taxonomy capturing common error patterns—such as temporal drift, symbol-grounding failure, and execution slips—and use it to reveal model-specific weaknesses across different architectures.
    
    \item \textbf{Empirical Insights on Robustness and Data Scaling:} 
    Through controlled experiments across specialist imitation and generalist VLA models, we uncover clear robustness–precision trade-offs and show that performance on long-horizon, deformable-object tasks saturates with limited demonstration scaling.
\end{itemize}

\section{Background}
Vision--Language--Action (VLA) models unify advances in computer vision, natural language processing, and robot learning. They leverage large pre-trained vision--language models (VLMs) and language models (LLMs) as the perception and reasoning front-end, attaching an action module that outputs low-level robot controls. Training uses triplets of visual observations, textual instructions, and action trajectories, enabling the model to predict actions directly from what it sees and what it is asked. This end-to-end paradigm blurs traditional boundaries between perception, planning, and control: instead of manually defined intermediate representations, the VLA implicitly maps semantic concepts to motor commands \cite{ahn2022saycan,driess2023palme,huang2023voxposer}. Inheriting broad world knowledge from VLMs/LLMs, VLA models exhibit strong generalization; for example, RT-2 can interpret unseen objects or instructions using its web-trained priors and attempt tasks never explicitly demonstrated. This ability to transfer semantic knowledge to robotic action distinguishes VLAs from earlier purely behavior-cloned policies.

We briefly survey the specific VLA models that we evaluate in this work, highlighting their design and prior results: 
\begin{itemize}
  \item 
  \textbf{ACT (2023)}---the \textit{Action Chunking Transformer} \cite{zhao2023act} is a task-specific imitation-learning policy for ALOHA, designed to overcome compounding errors in pixel-to-action control for fine manipulation. ACT uses a ResNet-18 vision encoder and a Transformer-based CVAE policy head that outputs short ``action chunks'' of future joint positions, reducing error accumulation and capturing the non-Markovian, multimodal nature of human demonstrations. A temporal-ensemble mechanism further stabilizes execution, enabling sub-millimeter bimanual manipulation from only $\sim$10 minutes of demonstrations. Despite its strong real-world performance, ACT is vision-only and single-task, lacking language conditioning or multi-task generalization. In our work, we therefore use ACT as a strong vision-only baseline to evaluate the benefits of language-conditioned manipulation models.

  \item 
  \textbf{OpenVLA--OFT} is a 7B open-source VLA model trained on nearly one million demonstrations from the Open X-Embodiment dataset \cite{openvla,openxembodiment}, integrating dual vision transformers (DINOv2, SigLIP) with a Llama-2 (7B) language backbone to form a unified multimodal manipulation policy. While OpenVLA exhibits strong task execution and semantic generalization, it requires fine-tuning for specific robots and tasks, and suffers from slow inference (3--5\,Hz) and instability in bimanual settings. The Optimized Fine-Tuning (OFT) method \cite{openvla_oft} addresses these limitations through parallel decoding, continuous action representation, and an L1 regression objective. Parallel decoding eliminates autoregressive action generation, enabling fast, synchronous control and boosting performance on LIBERO from 76.5\% to 97.1\%. On the real ALOHA bimanual platform, OFT achieves $\sim$15\% success rate and $\sim$43$\times$ faster inference than baseline OpenVLA.
  
  \item 
  \textbf{RDT-1B (2024)} is a 1B-parameter diffusion-based VLA model for bimanual manipulation \cite{rdt1b}. It integrates frozen SigLIP vision and T5-XXL language encoders, projected into a unified transformer token space, and outputs actions via an MLP head in a Physically Interpretable Unified Action Space. Pre-trained on 46 datasets with over 1M trajectories and fine-tuned on a 6k-trajectory dual-arm dataset, RDT-1B produces smooth, coordinated multi-step bimanual actions. It achieves strong real-world performance, including zero-shot transfer, accurate instruction following, and effective skill acquisition from 1--5 demonstrations.
  
  \item 
  \textbf{$\pi_0$ (2024)} \cite{pi0} is a general-purpose robotic foundation model built on a 3B-parameter PaliGemma-based vision--language backbone with a 300M action head. It introduces flow matching to align heterogeneous data and supports high-frequency control (up to 50\,Hz) across single-arm, dual-arm, and mobile platforms. $\pi_0$ follows a two-stage training regime: large-scale pre-training on the OXE superset (10{,}000+ hours, 68 tasks, 7 robot types), followed by task-specific fine-tuning. Experiments show strong generalization to new tasks and hardware, with rapid skill adaptation and performance surpassing prior VLA models.
  
  
\end{itemize}

\section{Related Work}
Research in robotic manipulation has traveled through different phases of development, from manual programming to task-specific control methods to general-purpose foundation models. In this section, we review prior research in four dimensions related to our evaluation framework: imitation learning, task-specific vision-action policies, vision--language--action foundation models, and benchmarking efforts in robotic manipulation.

\subsection{Imitation Learning for Robotic Manipulation}
Imitation learning (IL) enables robots to acquire behaviors from human demonstrations \cite{osa2018algorithmic}, with early milestones such as ALVINN \cite{pomerleau1989alvinn} and inverse reinforcement learning \cite{ng2000algorithms}. In manipulation, IL scales from pick-and-place to dexterous skills, yet suffers from limited generalization due to covariate shift: small errors drive the policy into unseen states, compounding failures \cite{ross2010efficient}. Approaches like DAGGER reduce this by collecting corrective data, but IL often requires large or carefully curated demonstrations \cite{robonet2019,walke2023bridgedata2}. Despite these limitations, IL remains fundamental, and modern VLA models still depend on expert trajectories for task- or robot-specific fine-tuning after pretraining.
learning control policies from examples or by inferring intent. In the context of manipulation, IL has been applied to tasks ranging from simple pick-and-place to complex dexterous manipulation. However, purely imitation-based policies often struggle to generalize beyond the conditions seen in their training demonstrations. A well-known problem is the covariate shift between training and execution: errors made by the learned policy can lead to states that were never observed in the expert data, causing the robot’s behavior to drift and compounding the error \cite{ross2010efficient}. Various improvements, such as DAGGER \cite{ross2010efficient}, have been proposed to mitigate this by intermixing corrective demonstrations during training. Nonetheless, IL methods typically require either large amounts of demonstration data or careful curation to cover expected variability \cite{robonet2019,walke2023bridgedata2}. Despite these challenges, IL remains central in robotics, and indeed, many recent VLA models still rely on IL datasets (expert trajectories) for fine-tuning their behavior on specific robots or tasks after a pretraining phase.

\noindent\textbf{Relation to our benchmark.}
Our evaluation uses IL-style dataset that include three cameras' data and dual-arm trajectories. ACT is trained from scratch on single task demonstrations; OpenVLA--OFT and RDT-1B are fine-tuned on the same ALOHA Mobile demonstrations; and $\pi_0$ is fine-tuned on single task demonstrations same as ACT policy. This lets us directly compare how IL-driven specialization versus IL-initialized foundation models transfer across our ID, Spatial OOD, and Instance{+}Spatial OOD evaluation settings.

\subsection{Task-Specific Vision--Action Policies}
Before large multimodal foundation models, robotic manipulation predominantly used \textbf{task-specific vision--action policies} that mapped perceptual inputs directly to low-level controls. Systems such as ACT \cite{zhao2023act} achieved sub-millimeter accuracy via carefully engineered imitation-learning pipelines, while RL and optimal-control variants leveraged structured policies and domain priors for speed and sample efficiency. These methods are highly precise within their training regimes but cannot generalize to unseen objects, language instructions, or new goals without retraining.

\noindent\textbf{Deformable-object manipulation.}
Cloth-like materials introduce non-rigid dynamics, occlusion, and tension-control challenges. Prior work includes \emph{FlingBot} for dynamic wrinkle removal \cite{ha2021flingbot}, grasp-and-pull policies via supervised learning \cite{seita2019fabric}, and \emph{SpeedFolding} for standardized bimanual folding routines \cite{avigal2022speedfolding}. These studies emphasize that success hinges on grasp-point selection, coordinated motion, and tension management rather than solely on object recognition.

\noindent\textbf{Relation to our benchmark.}
While these task-specific policies and domain-focused studies have produced reliable results in isolation, they lack a common experimental ground for systematic comparison across architectures, task types, and generalization conditions. Our benchmark extends this landscape by evaluating both task-specific and generalist VLA policies under a unified framework across real and simulated environments. Unlike prior single-task evaluations, our experience-driven study contrasts precision-oriented specialist models with large-scale VLA foundation models to reveal their respective trade-offs in accuracy, adaptability, and language grounding under controlled yet realistic conditions.

\subsection{Vision--Language--Action Foundation Models}
Large pre-trained vision and language backbones have enabled a new class of \textbf{Vision--Language--Action (VLA) models} that unify perception, reasoning, and control for general-purpose manipulation. Early systems such as the RT family \cite{brohan2023rt,brohan2023rt2} showed that pretraining on large cross-embodiment datasets yields policies that follow natural-language instructions and generalize to unseen objects, with RT-2 demonstrating zero-shot reasoning via inherited vision--language priors. Open-source efforts like \emph{OpenVLA} \cite{openvla}, along with efficient tuning methods such as OFT \cite{openvla_oft} and speculative decoding \cite{wang2025specvla}, further broadened accessibility. Related multimodal frameworks---e.g., \emph{SayCan} \cite{ahn2022saycan}, \emph{PaLM-E} \cite{driess2023palme}, \emph{VIMA} \cite{jiang2023vima}, and \emph{VoxPoser} \cite{huang2023voxposer}---explore grounding high-level reasoning into executable low-level actions.

More recent architectures trade off generalization for speed or contact-rich control. \emph{RDT-1B} \cite{rdt1b} employs diffusion-based trajectory prediction for smooth bimanual manipulation, while \emph{Octo} \cite{octo2024} uses lightweight transformer-diffusion networks suited for consumer GPUs. In contrast, $\pi_0$ (2024) uses a flow-matching action head for 50\,Hz real-time control, and \emph{SmolVLA} \cite{smolvla2025} offers a compact 450M flow-matching variant for efficient asynchronous inference. These models illustrate a spectrum from large-scale generalist VLAs to compact, low-latency controllers.

\noindent\textbf{Relation to our benchmark.}
Our benchmark systematically evaluates representative models from both ends of this spectrum---\emph{ACT} \cite{zhao2023act} as a specialist imitation policy and three open VLA representatives (\emph{OpenVLA--OFT}, \emph{RDT-1B}, and \boldmath{$\pi_0$})---under a unified task suite and experimental protocol. Unlike prior works that report results on isolated tasks or within simulation-only contexts, our experience-driven study directly contrasts these models in both real and simulated environments, quantifying their trade-offs in accuracy, efficiency, adaptability, and instruction following. This perspective allows us to contextualize how current foundation-model paradigms perform under realistic constraints.

\subsection{Benchmarks and Positioning of This Work}
Although robotic manipulation has advanced rapidly, systematic evaluation and benchmarking have not kept pace. Existing datasets and suites such as LIBERO \cite{liu2023libero}, OXE \cite{openxembodiment}, and the evaluation frameworks for RT-2 and PaLM-E \cite{colosseum2024,maniskill2,maniskill3,robocasa2024} primarily focus on simplified or simulation-only settings. For example, LIBERO emphasizes task transfer and contextual generalization, but most tasks involve single-arm pick-and-place operations or limited language grounding. These benchmarks are valuable for assessing specific capabilities such as semantic reasoning or cross-embodiment transfer, yet they overlook the challenges of real-world bimanual coordination, deformable-object handling, and robustness to out-of-distribution (OOD) variations. Recent surveys further emphasize the lack of standardized task definitions, unified evaluation protocols, and comparable metrics across robotic learning paradigms \cite{zhou2023survey,li2024simpler}. 

Building on these insights, our work provides a unified and experience-driven evaluation of representative models from both task-specific and foundation-model paradigms under identical protocols. Whereas prior benchmarks typically examine one architecture across multiple tasks or compare limited model variants within the same paradigm, we conduct a direct cross-paradigm comparison involving a specialist imitation-learning policy (ACT \cite{zhao2023act}) and three generalist VLA foundation models (OpenVLA--OFT \cite{openvla_oft}, RDT-1B \cite{rdt1b}, and $\pi_0$). The benchmark features four dual-arm, household-inspired manipulation tasks that integrate tool use, deformable-object control, spatial reasoning, and language understanding, all implemented on the standardized ALOHA Mobile platform \cite{zhao2023act}. By combining real-world and simulation-based evaluations, this study highlights how current foundation-model approaches compare with specialized policies in terms of precision, adaptability, and instruction grounding—offering a clearer view of the trade-offs shaping the next generation of robotic manipulation benchmarks.

\section{Unified Benchmark}
\label{sec:benchmark}

This section describes our evaluation frameworks for four different bimanual robot manipulation models on the ALOHA Mobile platform. We selected three foundation models (OpenVLA--OFT, RDT-1B, and $\pi_0$) and one task-specific model (ACT) for the evaluation. We also designed an additional evaluation on $\pi_0$ and OpenVLA--OFT in a designed LIBERO simulator \cite{liu2023libero} for further analysis. OpenVLA--OFT, RDT-1B, and $\pi_0$ are vision--language--action models that accept language instructions and were pretrained on large-scale datasets. In contrast, ACT is a vision-only imitation policy trained from scratch for each task. OpenVLA--OFT and RDT-1B are fine-tuned on the whole dataset, $\pi_0$ is fine-tuned on each single task, and ACT is trained from scratch on each single task. All the datasets we used are collected from our ALOHA mobile platform, and evaluations were conducted in a controlled environment using the same robot hardware and setup. 

\subsection{Physical Benchmark}
\label{subsec:phys-bench}

\subsubsection{\textbf{Setting}}
\label{subsec:physical-setting}
\begin{figure}[t]
\centering
\begin{tikzpicture}
\begin{axis}[
    xbar,
    width=0.9\linewidth,
    height=3cm,
    xmin=0, xmax=23,
    xlabel={Training time (days)},
    symbolic y coords={ACT,Pi0,OpenVLAOFT,RDT1B},
    ytick=data,
    yticklabels={ACT,$\pi_{0}$,OpenVLA,RDT-1B},
    y=4.5mm,                           
    bar width=7pt,
    enlarge y limits=0.4,            
    axis x line=bottom,
    tick align=outside,
    tick label style={font=\small},
    xlabel style={font=\small, yshift=4pt},
    xmajorgrids=true,
    grid style={dashed},
    nodes near coords,
    every node near coord/.append style={
        font=\small,
        anchor=west,
        xshift=2pt,
        text=blue!60
    },
    /pgf/number format/fixed,
    /pgf/number format/precision=2,
    clip=false                       
]
\addplot[fill=blue!15,draw=blue!60] coordinates
    {(0.17,ACT) (2,Pi0) (21,OpenVLAOFT) (21,RDT1B)};
\end{axis}
\end{tikzpicture}
\caption{Training time on our setup (days): ACT $\approx$0.17, $\pi_{0}$ $\approx$2, OpenVLA--OFT and RDT-1B $\approx$21 each.}
\Description{Horizontal bar chart with four rows showing training time in days: ACT 0.17, $\pi_{0}$ 2, OpenVLA--OFT 21, RDT-1B 21.}
\label{fig:training_time}
\end{figure}
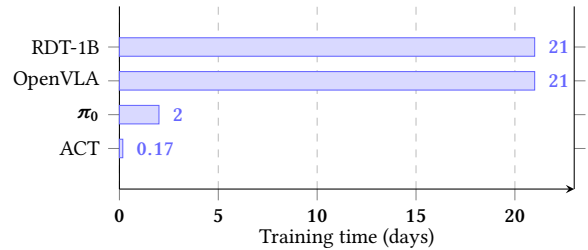

\paragraph{\textbf{Platform.}} The real-world hardware we are using is the ALOHA mobile platform \cite{zhao2023act}, which comprises two humanoid-sized robotic arms with seven degrees of freedom, positioned on a shared base in a shoulder-like structure. A parallel jaw gripper is attached to each robotic arm. The robot operates in front of a 1-meter by 1-meter workbench. To minimize visual distractions, we kept the overhead lighting constant and used a white tabletop background for all trials. 
\paragraph{\textbf{Perception.}} The perception device consists of three Intel RealSense D405 RGB-D cameras: one top-mounted camera covers the entire working area, and two wrist-mounted cameras are mounted on the end of the robot arm. All cameras send image data with a resolution of 640×480 in real time at a frame rate of 30\,Hz. Models that support multi-view input can simultaneously receive perceptual information from three camera views.
\paragraph{\textbf{Computation configuration}} 
All tests are conducted on an NVIDIA RTX 5090 GPU, with a single-step control time budget of \textbf{\SI{40}{ms} (25\,Hz)}---the allotted time for sensing, computation, and actuation in each control cycle.
OpenVLA--OFT always runs stably at 25\,Hz on the ALOHA platform and meets the time limit requirements. $\pi_0$ supports a maximum control frequency at 50\,Hz, which can also be satisfied. The ACT controller runs at 50\,Hz, also meeting the real-time requirements. OpenVLA--OFT and RDT-1B are fine-tuned on dual A6000 GPUs since they require large memories, and they have been trained for over 2 weeks. ACT and $\pi_0$ are trained on RTX5090 GPUs. RTX5090 can finish training the ACT model in 2 hours and $\pi_0$ in 24 hours, which is much faster than OpenVLA--OFT and RDT-1B, as seen in the Table~\ref{fig:training_time}. Their training parameters are shown in Table~\ref{tab:train_finetune_configs}. Table~\ref{tab:train_finetune_configs} summarizes the per-task configurations used to train or fine-tune each model.


\begin{table}[t]
\centering
\small
\setlength{\tabcolsep}{6pt}
\renewcommand{\arraystretch}{1.05}
\resizebox{\columnwidth}{!}{%
\begin{tabularx}{\linewidth}{@{}l c X c X@{}}
\toprule
\textbf{Model} & \textbf{Mode} & \textbf{Inputs} & \textbf{Batch / LR} & \textbf{Steps / Schedule} \\
\midrule
OpenVLA--OFT (7B) & FT      & 3 cams + proprio & 2 / $5\!\times\!10^{-4}$  & 160k; val/5k, ckpt/10k \\
RDT-1B            & FT      & 3 cams + proprio & 16 / $1\!\times\!10^{-5}$ & 200k; samp/1k, ckpt/2k \\
ACT               & Scratch & 3 cams + proprio & 2 / $1\!\times\!10^{-5}$  & 3k total \\
$\pi_{0}$         & FT      & 3 cams + proprio & 32 / $1\!\times\!10^{-5}$               & 30k; log/100, ckpt/1k \\
\bottomrule
\end{tabularx}}
\caption{Per-task training/fine-tuning (FT) settings.}
\label{tab:train_finetune_configs}
\end{table}

\subsubsection{\textbf{Tasks \& Evaluation Metrics}}
\label{subsec:physical-tasks} 

\begin{figure}[t]
    \centering
    \begin{subfigure}[b]{0.4\linewidth}
        \includegraphics[width=\linewidth]{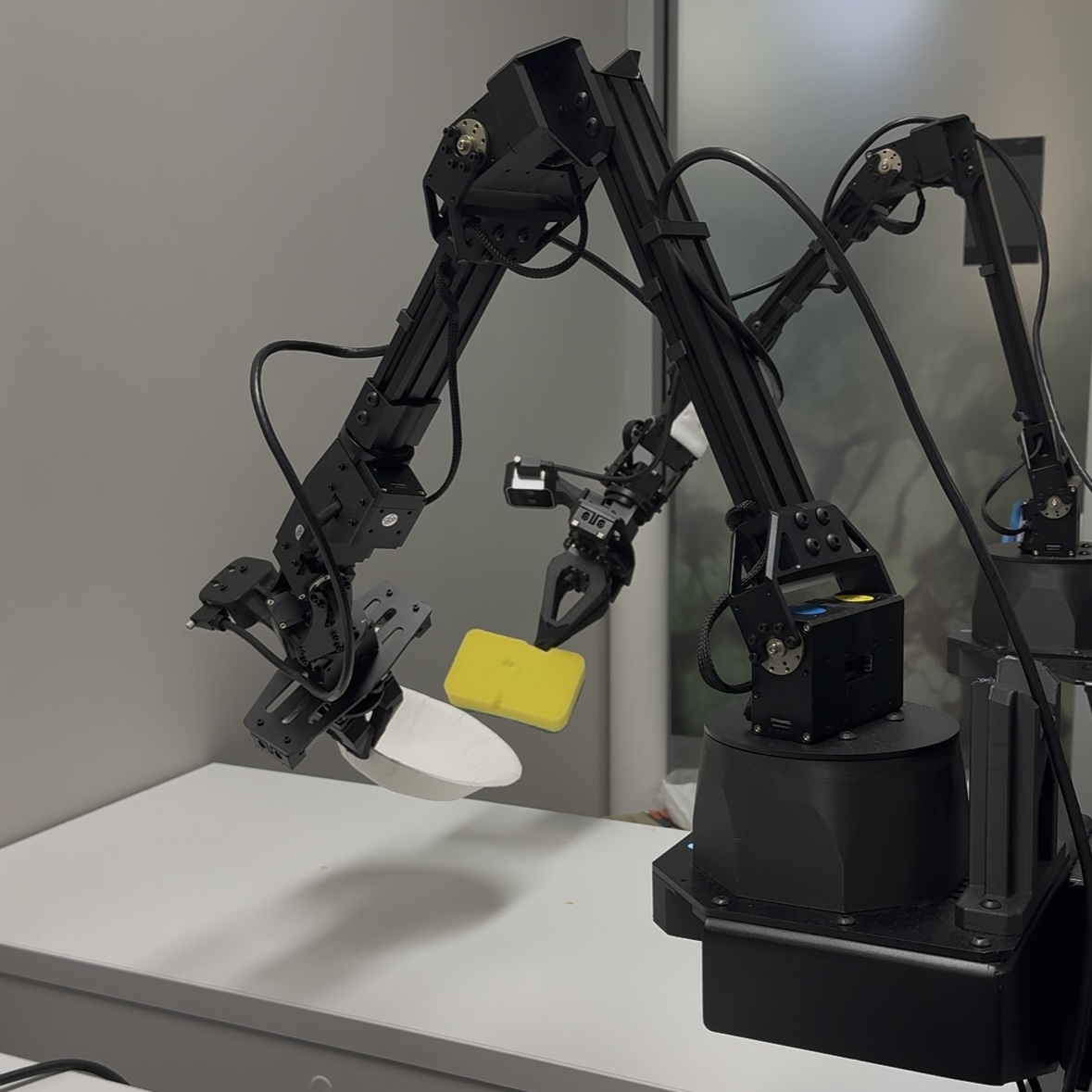}
        \caption{Washing Dish}
    \end{subfigure}
    \hfill
    \begin{subfigure}[b]{0.4\linewidth}
        \includegraphics[width=\linewidth]{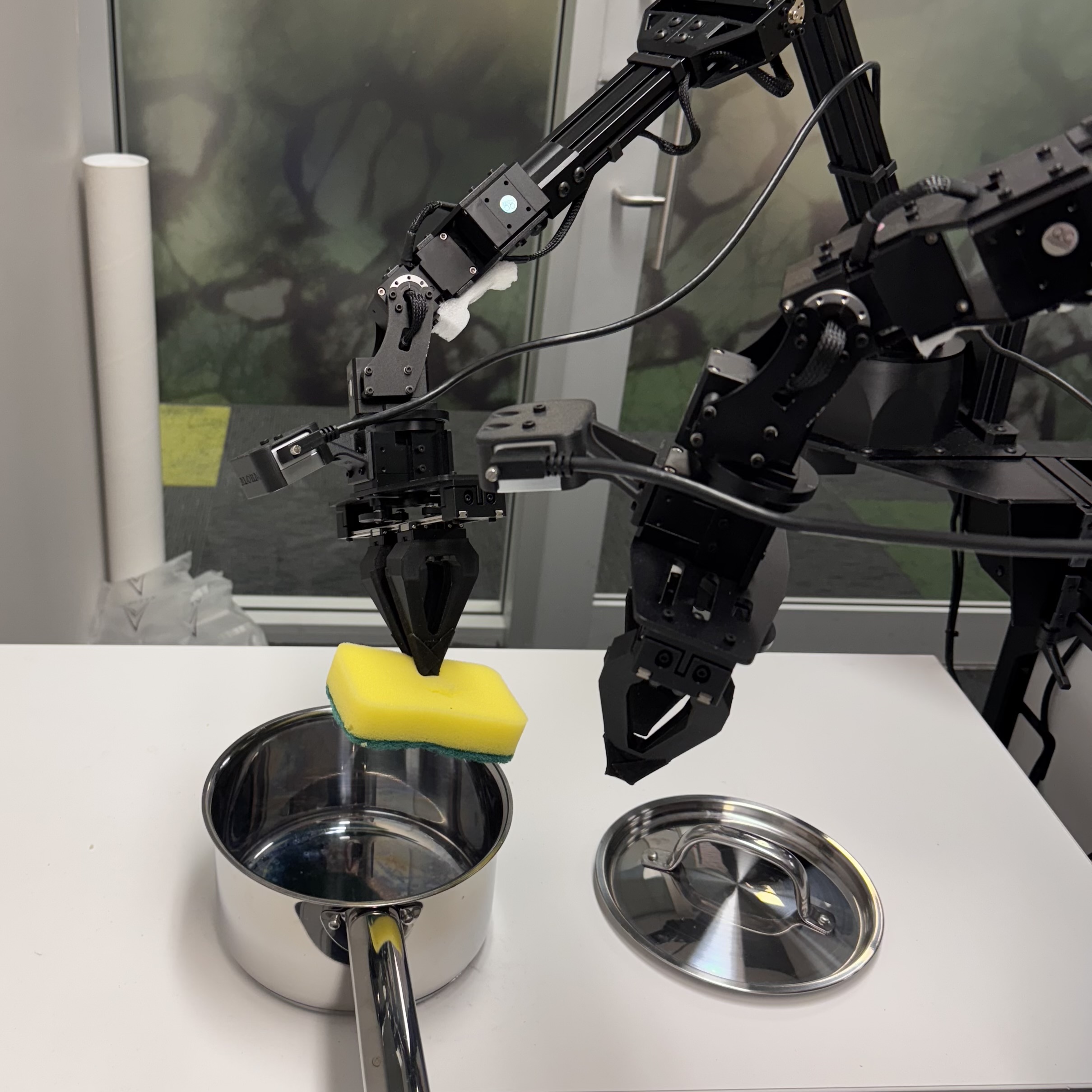}
        \caption{Put sponge into pot}
    \end{subfigure}

    \vspace{0.6em}

    \begin{subfigure}[b]{0.4\linewidth}
        \includegraphics[width=\linewidth]{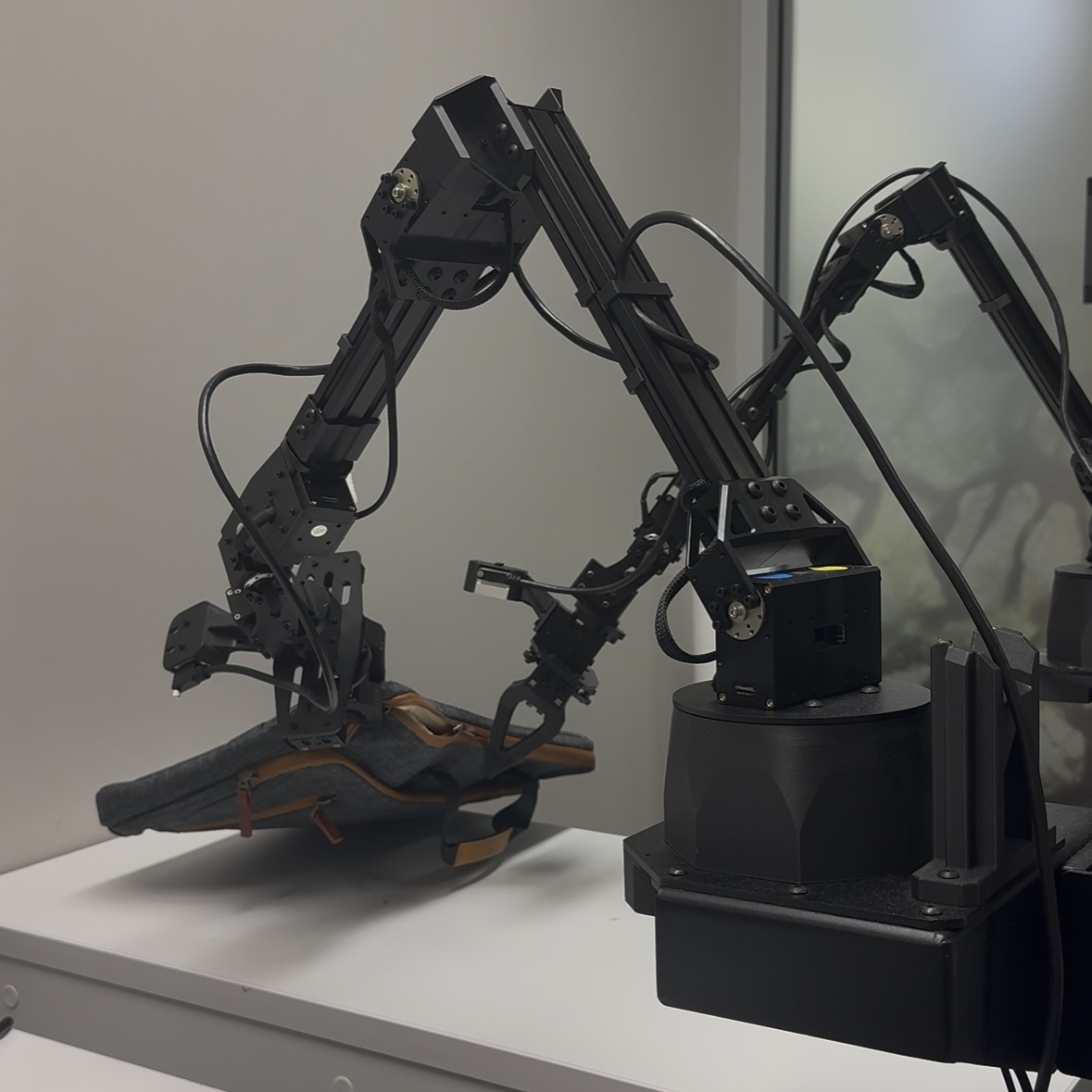}
        \caption{Unzip Bag}
    \end{subfigure}
    \hfill
    \begin{subfigure}[b]{0.4\linewidth}
        \includegraphics[width=\linewidth]{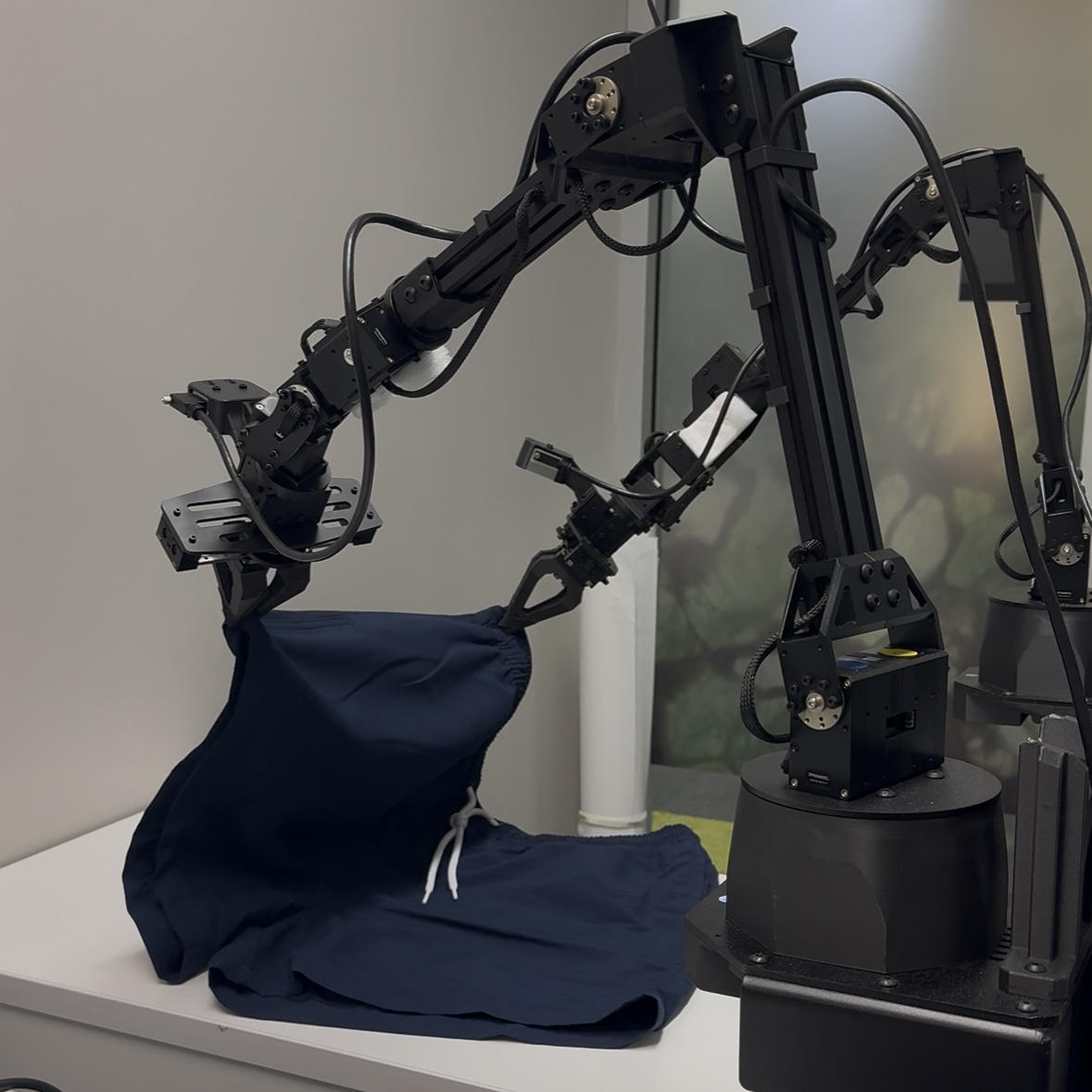}
        \caption{Folding Shorts}
    \end{subfigure}

    \caption{Real-world manipulation tasks on ALOHA Mobile.}
    \Description{Four photos on a white table: cleaning a dish with a sponge; placing a sponge into a pot; unzipping a soft laptop bag; folding shorts with two arms.}
    \label{fig:real-tasks}
\end{figure}

\paragraph{\textbf{Tasks.}}
We created four different manipulation tasks (illustrated in \Cref{fig:real-tasks}a–d) for the evaluation framework. These tasks include tool utilization, bimanual manipulation, instruction understanding, and deformable objects manipulation. All tasks involve coordinated bimanual manipulation and are presented to the robot as an instruction. 

\begin{enumerate}
\item \textbf{Clean Dish.} The robot is instructed to clean the dish using a sponge. The robot needs to use the left gripper to pick up the dish, lift it to the middle, and keep it stable. Use the right gripper to pick up the sponge to clean the surface of the dish. This task requires multiple steps to complete. The robot needs to coordinate the two grippers, with one arm holding the dish while the other uses a sponge to wipe it. Language commands used for this task are ``Wipe the plate with the sponge" and ``Clean the dish". This task tests the model's precise force control and bimanual coordination ability. The robot must apply enough pressure to clean the surface without breaking the paper plate. We evaluated the model's generalization ability by altering the starting positions of the objects. The key challenges for the model include using the sponge correctly, understanding the concept of ``clean" and bimanual manipulation. 

\item \textbf{Put Sponge into Pot.} This task aims to evaluate the model's object selection and spatial understanding capability. The experimental setup includes a table with a cooking pot with a removable lid. There are two other objects to test the object selection skills. The instructions explicitly describe the target object. For example: ``Put the sponge/donut in the pot". The steps are as follows: (i) remove the pot cover with the left gripper; (ii) pick up and hold the object with the right gripper; (iii) move the right gripper on top of the pot, then release the gripper. We investigated two situations. In-distribution refers to the scenario where the target object and the position of the object stay the same as in the dataset. Out-of-distribution refers to the use of unseen objects and moving the position of the objects randomly. For example, the training object is using ``sponge", while the testing object was a ``yellow donut" toy. A trial is considered successful if the robot puts the correct item in the pot. This task evaluates the model's object recognition skills, spatial understanding, and multi-step planning skills.

\item \textbf{Unzip Bag.} This task aims to evaluate the model's manipulation of a deformable object (bag is using soft material), precise manipulation (zipper is a tiny object), and bimanual manipulation. The commands we used are ``Unzip the laptop bag", ``Open the bag.", and ``Use left gripper to lift and secure the bag, use right gripper to grasp the zipper and slide to open the bag". The steps required to consider success of the task are: (i) Hold one edge of the bag steady with one gripper to create tension. (ii) Pinch the zipper tab or slider with the other gripper. (iii) Pull the slider smoothly from one end stop to the other to fully open the zipper. The primary challenge is maintaining a straight, continuous pull while keeping the fabric taut to avoid snags or partial openings. For generalization testing, we vary bag orientation and color or material. We also add mild occlusions requiring a small reposition to grasp the tab. To be successful, the zipper moved to another end, and the bag is open, and it is not damaged. 

\item \textbf{Folding Shorts.} This task evaluates bimanual coordination on a deformable object. A pair of shorts lies flat on the table. The instruction is: ``Fold the shorts twice" or ``Fold the shorts". The required steps are as follows: (i) Use two separate grippers to hold the left and right waistband edges. (ii) Lift and bring them together so the shorts fold roughly in half. (iii) With the right gripper, grab the waistband and fold it toward the leg until the edges meet, while the left gripper stabilizes the legs or midpoint to prevent dragging. Challenges include maintaining the cloth's shape during regrasp and avoiding partial lifts or bunching. Success means two-fold along perpendicular axes to make a quarter-size rectangle. Failure settings include garments falling off, folding twice along the same axis, or forming a crumpled mess. We train the VLA models on 100 and 200 demonstrations, comparing their performance to isolate the effect of data volume. We also use shorts with different colors to evaluate the models' generalization capability. 
\end{enumerate}

All the training data are collected via teleoperation. For each task, demonstrations covered a diverse range of initial object positions and orientations to improve the data quality. For language-grounding tasks, we varied the phrasing of instructions in the demonstrations to improve VLA models' language understanding capability. For example, the ``Clean Dish" task uses both the command ``Clean Dish" and the more detailed one: ``use right gripper to grasp the sponge, use left gripper to grasp the dish, and use sponge to clean the dish." We recorded 100 demonstrations on the ``Clean Dish", ``Put Sponge into Pot", and ``Unzip Bag" tasks. We collected 200 demonstrations on the ``Folding Shorts" task. 

\paragraph{\textbf{Evaluation Metrics.}}

We evaluate under the three \emph{evaluation settings} defined in the Introduction (ID, Spatial OOD, and Instance{+}Spatial OOD). Unless noted otherwise, we run $n{=}50$ trials per (task, model, setting). A trial terminates upon success, timeout/horizon, or safety stop (the latter two count as failures). For OOD settings, we uniformly randomize object poses; for Instance{+}Spatial OOD, we additionally replace objects with unseen instances. To ensure fairness, we reuse identical randomized initial states across models (shared seeds) and interleave evaluations in a round-robin fashion.

We assess performance using three metrics:
\begin{enumerate}
  \item \textbf{Task Success Rate (SR):} percentage of trials that achieve the task goal, indicating overall sequence execution success.
  \item \textbf{Time to Success (TTS):} elapsed time from the first action to task completion, measured only on successful trials (lower is better).
  \item \textbf{Instruction Adherence Accuracy (IAA):} per-command correctness of target and relational grounding for language-specified goals.
\end{enumerate}
We report Wilson-score 95\% confidence intervals for SR and IAA.

\begin{figure}[!htbp]
  \centering
  \includegraphics[width=0.3\textwidth]{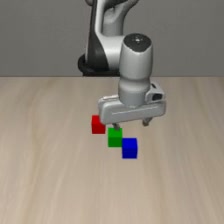}
  \caption{Simulation snapshot showing the robot arm manipulating colored cubes in MuJoCo.}
  \label{fig:sim-setup}
\end{figure}

\subsection{Simulation Benchmark}
\label{subsec:sim-bench}

To complement the real-robot evaluations, we designed a controlled \emph{simulation benchmark} to assess zero-shot language grounding and spatial reasoning, as illustrated in Figure~\ref{fig:sim-setup}. The benchmark uses a simple MuJoCo tabletop environment with three colored cubes (red, green, and blue) randomly arranged on the table. Tasks in this setup evaluate the model’s ability to understand and execute natural language instructions that involve spatial relations and sequencing.

\subsubsection{\textbf{Settings}}
\label{subsec:sim-setting}

We use a MuJoCo-based tabletop environment containing three cubes (red, green, and blue). Evaluations are conducted using \textbf{OpenVLA--OFT} and \textbf{$\pi_0$}, as they are the only models with publicly available checkpoints compatible with the LIBERO dataset. 

Two evaluation conditions are used:
\begin{itemize}[leftmargin=*]
    \item \textbf{In-distribution (ID):} cube positions remain fixed at the center of the table, reflecting configurations seen during training.
    \item \textbf{Out-of-distribution (OOD):} cube positions are randomized in each trial to test spatial generalization.
\end{itemize}

Each model is evaluated for $n{=}50$ trials per (task, setting) using identical random seeds and a round-robin evaluation order for fairness. A trial terminates upon task success, timeout, or safety stop.

\subsubsection{\textbf{Tasks \& Evaluation Metric}}
\label{subsec:sim-tasks}

\paragraph{\textbf{Tasks.}} 
Two representative manipulation tasks are used to probe language grounding and spatial reasoning:

\begin{enumerate}[leftmargin=*]
    \item \textbf{Pick by Color:} The agent must pick up the cube corresponding to the specified color (e.g., ``Pick up the red cube''), evaluating color grounding and instruction following.
    \item \textbf{Stack by Order:} The agent is instructed to stack the cubes in a given order (e.g., ``Stack the cubes in the order red, green, blue''), assessing sequence understanding and spatial composition.
\end{enumerate}

\paragraph{\textbf{Evaluation Metric.}} 
The only metric considered is the \textbf{Task Success Rate (SR)}, defined as the percentage of trials where the commanded task goal is achieved. For stacking, success additionally requires the final structure to be physically stable and correctly ordered.

\section{Benchmark Results: Real World}
\label{sec:real-results}\label{subsec:aloha_results}

\subsection{Overall Performance}
\begin{table}[t]
\centering
\caption{Macro-averaged success rates (\%) across evaluation settings.}
\label{tab:macro-sr}
\setlength{\tabcolsep}{4pt}
\small
\resizebox{\linewidth}{!}{%
\begin{tabular}{lccccc}
\toprule
\textbf{Model} & \textbf{ID SR} & \textbf{Spatial OOD (seen) SR} & $\boldsymbol{\Delta}$ \textbf{vs ID} & \textbf{Instance{+}Spatial OOD SR} & $\boldsymbol{\Delta}$ \textbf{vs ID} \\
\midrule
$\pi_0$    & 72.3\% & 66.8\% & $-5.5$  & 58.5\% & $-13.8$ \\
ACT        & 48.3\% & 17.8\% & $-30.5$ &  5.5\% & $-42.8$ \\
RDT-1B     & 13.5\% &  6.8\% & $-6.7$  &  6.3\% & $-7.2$  \\
OpenVLA    & 13.0\% &  6.5\% & $-6.5$  &  6.0\% & $-7.0$  \\
\bottomrule
\end{tabular}}
\\[3pt]
\footnotesize \emph{Note:} Macro-average is the unweighted mean over the four core tasks \{Clean Dish, Put Sponge in Pot, Unzip Bag, Folding Shorts (200)\} to avoid double-counting the folding task. 
\end{table}

\Cref{tab:macro-sr} reports the \textit{macro-averaged success rates} across the three evaluation settings (ID, Spatial OOD, and Instance{+}Spatial OOD). The term \textit{macro-averaged} refers to the unweighted mean success rate computed over the four core tasks in our benchmark (clean dish, put sponge in pot, unzip bag, and fold shorts). 

From these results, $\pi_0$ achieves the highest overall success rate across all experiments and demonstrates the strongest robustness to distribution shifts. In the standard (ID) setting, $\pi_0$ attains an average success rate of about 72\%, which decreases only by roughly 5\% under spatial perturbations and by 13\% when evaluated on unseen objects. 

In comparison, ACT reaches 48\% in the standard setting but drops by about 30\% under spatial randomization and achieves only 5\% in unseen object scenarios. The two VLA foundation models, OpenVLA and RDT-1B, achieve low success rates (around 13\%) on in-distribution tasks, further decreasing to 6--7\% under out-of-distribution conditions. 

Overall, these results indicate that $\pi_0$ not only outperforms all baseline models but also generalizes more robustly to new environments and objects. Specifically, $\pi_0$ loses about 14 percentage points (a relative decrease of 19\%) between standard and unseen-object settings, whereas ACT loses over 42 percentage points (88\% relative drop), and OpenVLA/RDT-1B lose about 7 points (approximately 50\%) from an already low baseline. This demonstrates that $\pi_0$ generalizes significantly better, while ACT remains brittle under distribution shifts, and pre-trained base models without sufficient adaptation fail to reach a strong performance baseline.

\subsection{In-Distribution Setting (Training Conditions)}
Table~\ref{tab:id} summarizes the performance of each model on the four tasks under the in-distribution (ID) condition, which corresponds to the scenarios present in the training data (i.e., seen objects in nominal placements). We report success rates (SR) and average time to success (TTS) for each model–task pair. An em-dash (—) for TTS indicates that there were no successful trials, and thus TTS is undefined. In this setting, which closely matches the training distribution, models are expected to achieve their highest performance.

\label{subsec:results_id}
\begin{table}[H]
    \centering
    \caption{ID performance (seen objects; 50 trials per task per model). Em-dash: no successes (TTS undefined).}
    \label{tab:id}
    \begin{tabular}{lccccc}
        \toprule
        Task                  & Metric & ACT   & OpenVLA  & RDT-1B   & $\pi_0$  \\
        \midrule
        Clean Dish            & SR     & 86\%  & 29\%     & 27\%     & 87\%      \\
                              & TTS    & 13\,s & 24\,s    & 35\,s    & 15\,s     \\
        \hline
        Put Sponge in Pot     & SR     & 85\%  & 23\%     & 24\%     & 83\%      \\
                              & TTS    & 16\,s & 30\,s    & 85\,s    & 21\,s     \\
        \hline
        Unzip Bag             & SR     & 22\%  & 0\%      & 3\%      & 67\%      \\
                              & TTS    & 16\,s & \textemdash & 83\,s & 25\,s     \\
        \hline
        Folding Shorts (100)  & SR     & 0\%   & 0\%      & 0\%      & 0\%       \\
                              & TTS    & \textemdash & \textemdash & \textemdash & \textemdash \\
        \hline
        Folding Shorts (200)  & SR     & 0\%   & 0\%      & 0\%      & 52\%      \\
                              & TTS    & \textemdash & \textemdash & \textemdash & 70\,s \\
        \bottomrule
    \end{tabular}
\end{table}

Under the standard conditions, $\pi_0$ and ACT perform well on the ``Clean Dish" and ``Put Sponge in Pot" tasks, ACT and $\pi_0$ achieved success rates of 86\% and 85\%. These two tasks are relatively simple compared to the ``Folding Shorts" and ``Unzip Bag" tasks. However, OpenVLA--OFT and RDT-1B perform poorly even on these simple tasks, their performance are under 30\% SR. This result indicates that ACT and $\pi_0$ learned these simple tasks effectively from the dataset. In contrast, OpenVLA--OFT and RDT-1B achieved success rates around 23–29\% on the same tasks, suggesting that they didn't learned the action, vision, and language pattern efficiently. 

The ``Unzip Bag" task is more challenging compared to the ``Clean Dish" and ``Put Sponge in Pot" tasks. $\pi_0$ achieves a 67\% success rate, indicating that it has learned a reasonable policy, whereas ACT only reaches a 22\% success rate. OpenVLA--OFT and RDT-1B perform near-zero success on this task. The performance shows that $\pi_0$ outperforms the others on this more complex task. This task requires fine-grained coordination and handling of deformable parts, $\pi_0$ performed better than the
other two VLA models, possibly due to its broader pre-training and flow-matching strategy. We discovered the poor performance on the others is due to the misgrasp of the small zipper. 

All methods fail to complete the Folding Shorts task with a 100-episode training dataset, indicating that tasks with deformable objects and long-horizon tasks are challenging for the current policies. However, when we use the 200-episode ``Folding Shorts" dataset to train the models, $\pi_0$ achieves a 52\% success rate, whereas ACT, OpenVLA, and RDT-1B remain at 0\% success. The result from this task indicates that the quantity of the dataset has a huge impact on the complicated tasks. In summary, under the standard conditions, $\pi_0$ emerges as a top performer on both simple and complex tasks, ACT excels on the pick-place and bimanual coordination tasks, but fails on the tasks requires to handle deformable objects, and the two baseline VLA models underperform across all tasks.

\subsection{Spatial OOD with Seen Objects}
\label{subsec:results_spatial_seen}

We next examine the models’ performance when faced with spatial OOD settings using the same objects as the training dataset. The identities and appearance of objects remain same, but their positions and orientations on the table are randomized in each trial. This tests each model’s ability to generalize its learned skills to new spatial arrangements. Table~\ref{tab:spatial-ood} reports the results. 

\begin{table}[H]
    \centering
    \caption{Spatial OOD with seen objects; 50 trials each.}
    \label{tab:spatial-ood}
    \begin{tabular}{lccccc}
        \toprule
        Task                  & Metric & ACT      & OpenVLA     & RDT-1B     & $\pi_0$ \\
        \midrule
        Clean Dish            & SR     & 36\%     & 15\%        & 14\%       & 85\%     \\
                              & TTS    & 13\,s    & 27\,s       & 36\,s      & 16\,s    \\
        \hline
        Put Sponge in Pot     & SR     & 32\%     & 11\%        & 12\%       & 80\%     \\
                              & TTS    & 16\,s    & 31\,s       & 87\,s      & 22\,s    \\
        \hline
        Unzip Bag             & SR     & 3\%      & 0\%         & 1\%        & 56\%     \\
                              & TTS    & 16\,s    & \textemdash & 83\,s      & 25\,s    \\
        \hline
        Folding Shorts (100)  & SR     & 0\%      & 0\%         & 0\%        & 0\%      \\
                              & TTS    & \textemdash & \textemdash & \textemdash & \textemdash \\
        \hline
        Folding Shorts (200)  & SR     & 0\%      & 0\%         & 0\%        & 46\%     \\
                              & TTS    & \textemdash & \textemdash & \textemdash & 70\,s \\
        \bottomrule
    \end{tabular}
\end{table}

All the policies' performance decreases; $\pi_0$ has the smallest drop. ACT’s success rates drop sharply compared to the ID setup. For the \textit{Clean Dish} task, ACT falls from 86\% to only 36\% under OOD settings. For \textit{Put into Pot}, the percentage decreases from 85\% to 32\%. These represent roughly 50- 53\% drops for ACT on relatively simple tasks. This indicates ACT has poor spatial generalization capability. On the \textit{Unzip Bag} task, ACT's success rate drops from 22\% to just 3\%. The right gripper is unable to locate the zipper correctly, indicating that ACT policy is overfitting to the positions seen in demonstrations and locating small objects is challenging. 

OpenVLA and RDT-1B also suffer performance drops with spatial OOD, although given their low starting points, the absolute changes are smaller. For example, OpenVLA’s success on Clean Dish goes from 29\% to 15\%, approximately a 14\% drop, and RDT-1B’s Clean Dish goes from 27\% to 14\%. Both models end up below 15\% on Clean Dish and around 10–12\% on Put Sponge into Pot. We discovered that for OpenVLA--OFT and RDT-1B 80\% of the time, the gripper is able to move to the relatively correct location, which means they do have spatial understanding, but they fail the task due to small misalignment on the grasp step. For example, in the ``Put Sponge in Pot" task, the arm is able to move close to the sponge, but the gripper struggles to grasp the object due to a misalignment of less than 1cm. 

In contrast, $\pi_0$ maintains a very high level of performance despite the randomization of object positions. $\pi_0$ ’s success rate on the Clean Dish task is 85\% under spatial OOD, a tiny performance drop from 87\% compare to the ID settings. Similarly, for the “Put Sponge into Pot” task, it achieves 80\% (vs. 83\% in ID). Even on the harder Unzip Bag task, $\pi_0$ manages to succeed in 58\% success rate, only an 11\% drop from the ID scenario.

Turning to the Folding Shorts task, the story remains similar to the ID settings. All the models achieve a zero success rate with 100 episodes of training data. With a 200-episode training dataset, $\pi_0$ achieves a 46\% success rate. This is only slightly lower than its 52\% in the ID settings.

\subsection{Instance{+}Spatial OOD with Unseen Objects}
\label{subsec:results_unseen}

Finally, we evaluate the most challenging evaluation settings: Instance{+}Spatial OOD, in which the models must generalize to novel object instances (objects not encountered during training or fine-tuning) and to randomized spatial configurations simultaneously. Table~\ref{tab:inst-spatial-ood} presents the results for this setting, including SR and TTS for each model and task when presented with unseen objects in random positions.

\begin{table}[H]
    \centering
    \caption{Instance{+}Spatial OOD (unseen instances; random placements; 50 trials each).}
    \label{tab:inst-spatial-ood}
    \begin{tabular}{lccccc}
        \toprule
        Task                 & Metric & ACT   & OpenVLA & RDT-1B & $\pi_0$ \\
        \midrule
        Clean Dish           & SR     & 12\%  & 13\%    & 12\%   & 81\% \\
                             & TTS    & 17\,s & 26\,s   & 54\,s  & 34\,s \\
        \hline
        Put Sponge in Pot    & SR     & 10\%  & 11\%    & 12\%   & 73\% \\
                             & TTS    & 18\,s & 31\,s   & 90\,s  & 21\,s \\
        \hline
        Unzip Bag            & SR     & 0\%   & 0\%     & 1\%    & 37\% \\
                             & TTS    & \textemdash & \textemdash & 83\,s & 25\,s \\
        \hline
        Folding Shorts (200) & SR     & 0\%   & 0\%     & 0\%    & 43\% \\
                             & TTS    & \textemdash & \textemdash & \textemdash & 73\,s \\
        \bottomrule
    \end{tabular}
\end{table}

Under the Spatial OOD with Unseen Objects scenario, the performance gap grows notably wider. ACT’s policy, already challenged in spatial OOD settings, collapses completely under combined object and spatial OOD. For example, ACT’s success rate on Clean Dish drops from 86\% in Standard to approximately 12\% with unseen dishes. Similarly, ACT achieves just 10\% accuracy on ``Put into Pot" and 0\% accuracy on ``Unzip Bag" with unseen objects. The handful of ACT successes on Clean Dish is likely due to rare, fortunate alignments or trivially simple cases. This stability indicates that, compared to policies like ACT, which don’t have a pre-training strategy, the VLA foundation models exhibit stronger object understanding and robustness to rearrangements and novel instances.

Across both spatial OOD with seen objects and (ii) and instance{+}spatial OOD with unseen objects evaluations, the success rates of the two VLA baselines (OpenVLA and RDT- 1B) remain essentially unchanged. On Clean Dish, performance decreases by only 1–2\% success rate. This stability indicates that, compared to policies like ACT, which don’t have a pre-training strategy, the VLA foundation models exhibit stronger object understanding and robustness to rearrangements and novel instances.

The performance of $\pi_0$ stands with high success rates across all tasks, even under instance{+}spatial OOD settings. In the Clean Dishes task, $\pi_0$ achieved an 81\% success rate, only slightly below its 85\% success rate under ID settings. In the ``Put Sponge in Pot" task, $\pi_0$ achieved a 73\% success rate under spatial OOD settings, again only a modest drop from the 80\%+ success rate in ID settings. This instance{+}spatial OOD scenario evaluation results indicate that $\pi_0$ is able to generalize to objects of different appearance or shape, even with a different category. For example, even when given an unseen object, a donut toy, and asked $\pi_0$ to put it in a bowl, $\pi_0$ is still able to identify it and perform the task successfully over 50\%. In the more challenging Unzip Bag and fold shorts tasks, $\pi_0$ success rate achieved approximately 40\%, still the highest success rate compared to the other models.
\subsection{Case Study Analysis of Failure Modes}

This section presents a detailed case study analysis of representative failures observed during both real-world and simulation experiments. Rather than presenting a general taxonomy (which is discussed separately in Section~\ref{sec:failure-taxonomy}), we focus here on examining the concrete \emph{failure processes}, identifying their underlying causes, and linking them to corresponding recorded trials.

\begin{figure*}[t]
\centering
\begin{subfigure}{0.2\linewidth}
  \includegraphics[width=\linewidth]{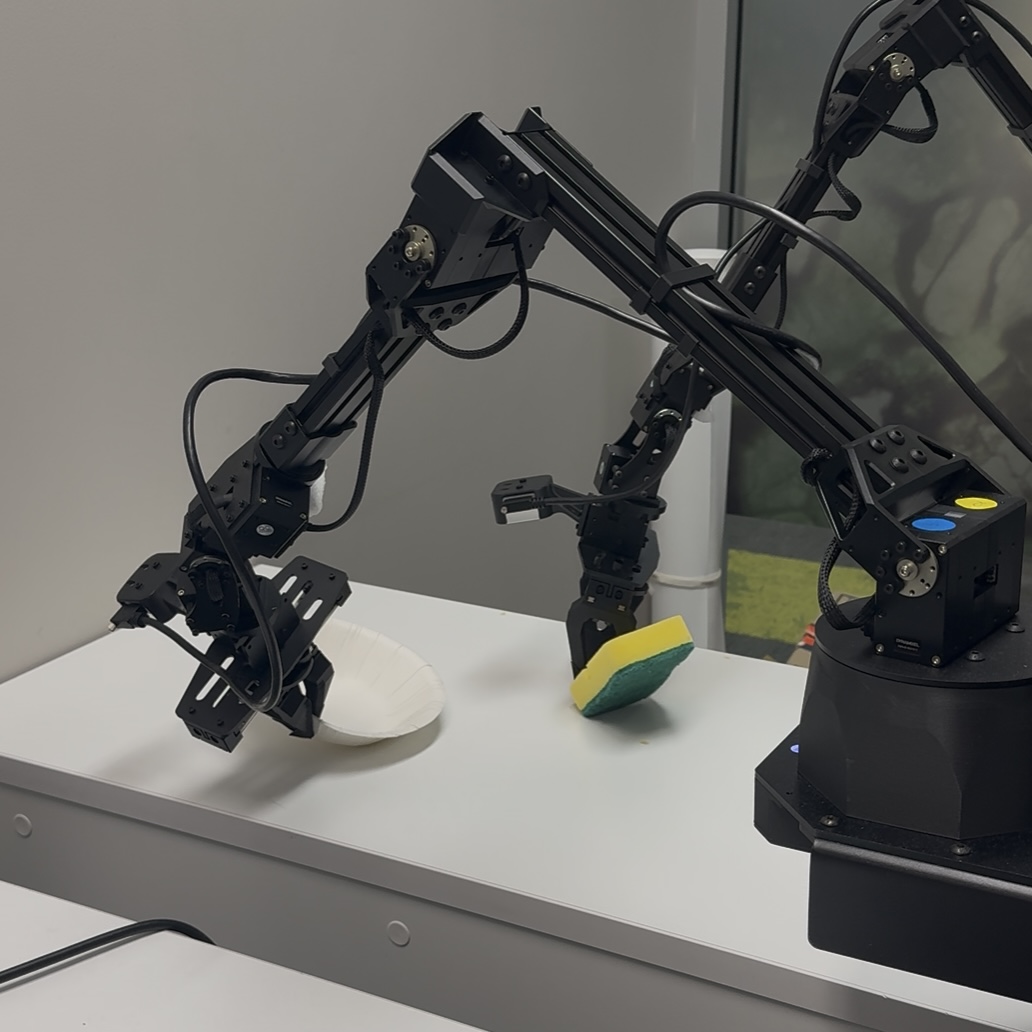}
  \caption{Clean Dish — planar (XY) miss-grasp on sponge (\textbf{Pre-grasp/Grasp}).}
  \Description{XY mis-grasp on the sponge; fingers pinch air or slip.}
\end{subfigure}\hfill
\begin{subfigure}{0.2\linewidth}
  \includegraphics[width=\linewidth]{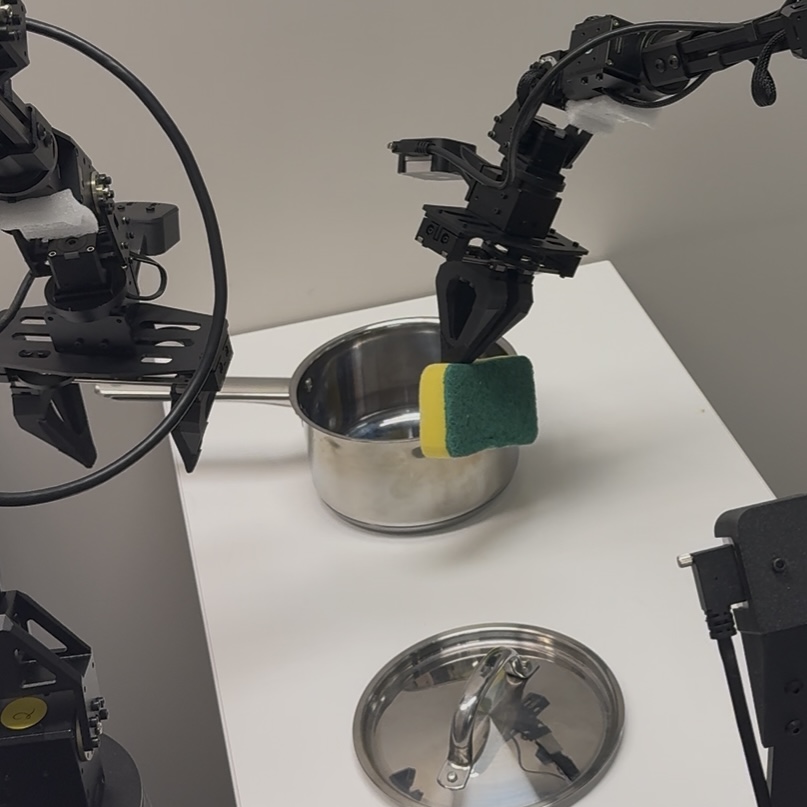}
  \caption{Put Sponge in Pot — small Z-offset; gripper not lifted high enough (\textbf{Pre-grasp/Grasp}).}
  \Description{Z undershoot or overshoot; contact too shallow or too deep.}
\end{subfigure}\hfill
\begin{subfigure}{0.2\linewidth}
  \includegraphics[width=\linewidth]{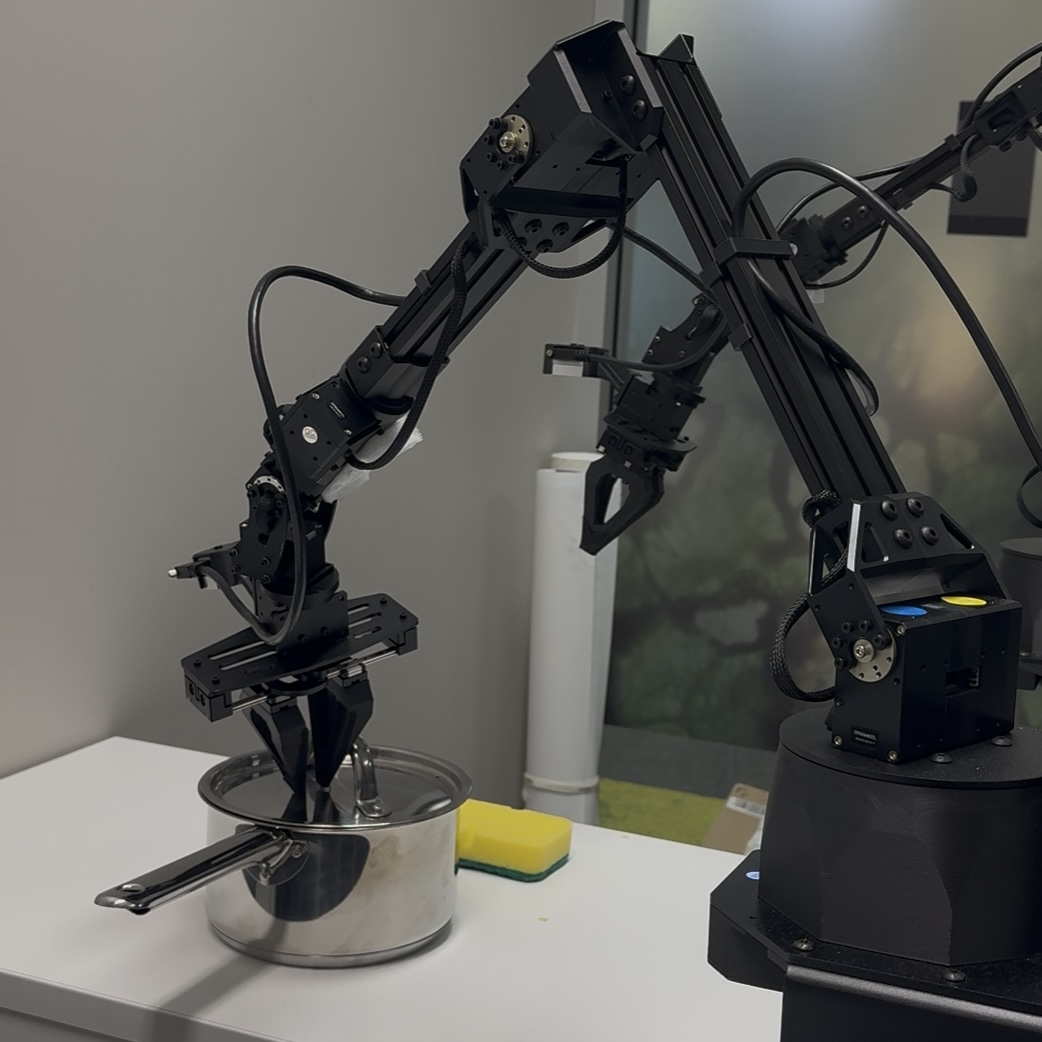}
  \caption{Put Sponge in Pot — lid-handle mis-grasp due to wrist yaw/pitch error (\textbf{Pre-grasp/Grasp}).}
  \Description{Approach orientation error; handle not securely pinched.}
\end{subfigure}\hfill
\begin{subfigure}{0.2\linewidth}
  \includegraphics[width=\linewidth]{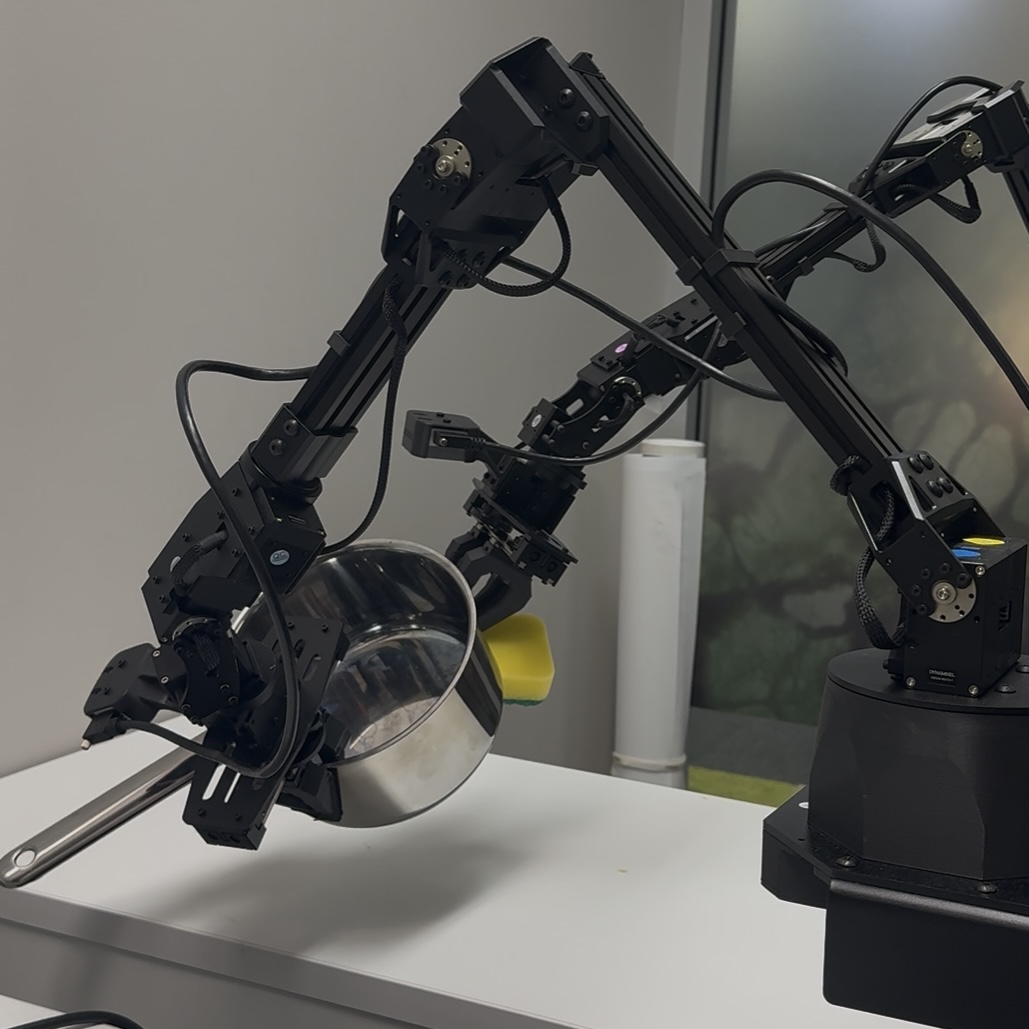}
  \caption{Put Sponge in Pot — instruction mis-follow; policy executes \emph{Clean Dish} instead (\textbf{Instruction adherence}).}
  \Description{Command is ``put sponge in pot'', but the policy executes the \emph{Clean Dish} routine, reflecting dataset-biased instruction grounding.}
\end{subfigure}

\vspace{0.6em}

\begin{subfigure}{0.2\linewidth}
  \includegraphics[width=\linewidth]{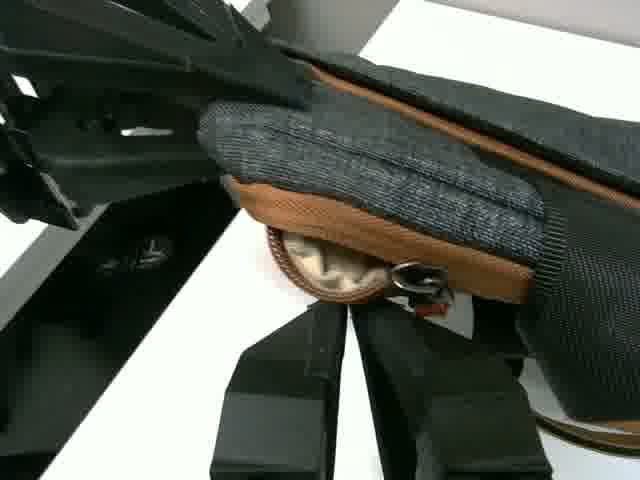}
  \caption{Unzip Bag — fails to grasp zipper tab; grasps fabric instead (\textbf{Pre-grasp/Grasp}).}
  \Description{Tiny target; finger–tab normal misalignment.}
\end{subfigure}\hfill
\begin{subfigure}{0.2\linewidth}
  \includegraphics[width=\linewidth]{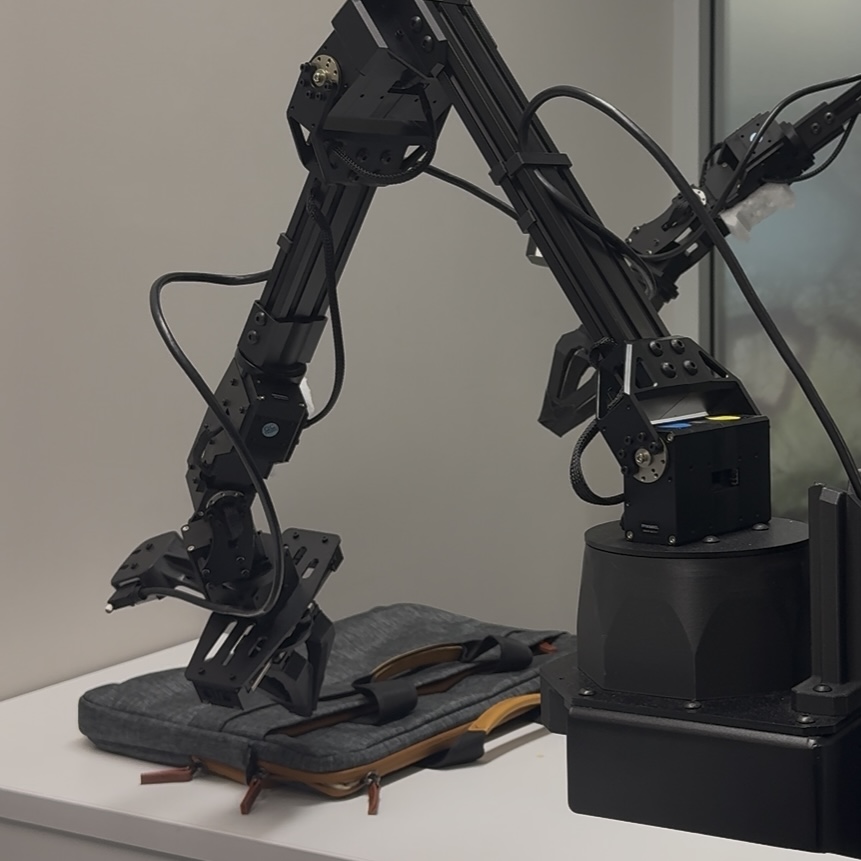}
      \caption{Pre-grasp---height (Z-axis) misalignment on bag (closes above surface)}
  \Description{Orientation drift causes incomplete opening.}
\end{subfigure}\hfill
\begin{subfigure}{0.2\linewidth}
  \includegraphics[width=\linewidth]{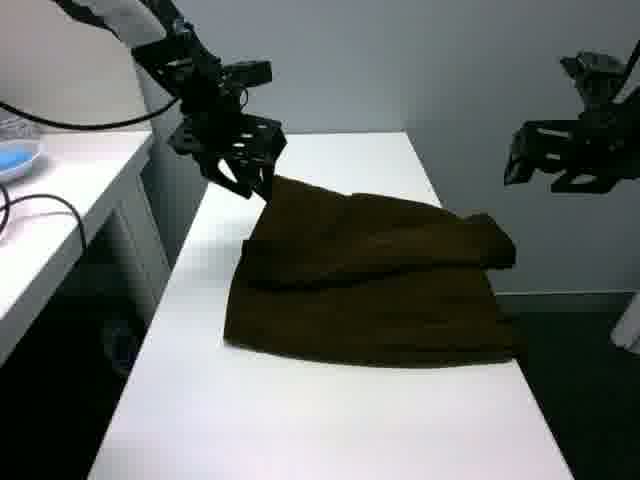}
  \caption{Folding Shorts — imperfect first fold (not fully in half); subsequent pose drift under occlusion (\textbf{Trajectory/State drift}).}
  \Description{Initial fold leaves the shorts uneven instead of halved; occlusion then causes accumulated state error that misaligns the second fold.}
\end{subfigure}\hfill
\begin{subfigure}{0.2\linewidth}
  \includegraphics[width=\linewidth]{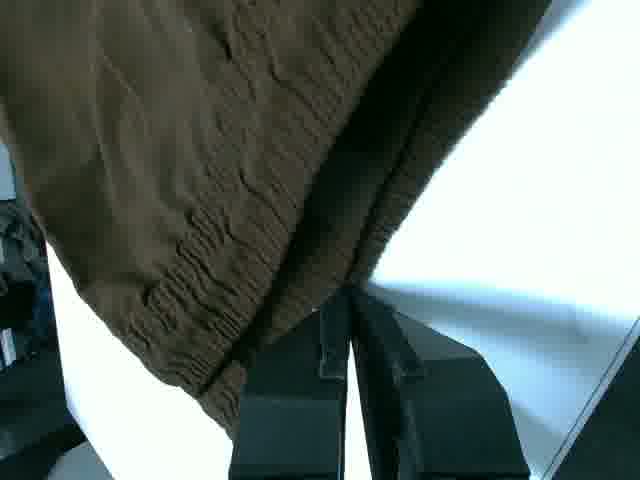}
  \caption{Folding Shorts — regrasp on waistband misses (\textbf{Pre-grasp/Grasp}).}
  \Description{Regrasp near small features; precision-limited.}
\end{subfigure}
\caption{\textbf{Real-world case studies aligned with our failure taxonomy (a–h).}
Many scenes are difficult to attribute to a single root cause. Ambiguities between perception precision, control timing, and state estimation make root-cause diagnosis non-trivial. We therefore complement qualitative video review with structured debugging (Sec.~\ref{subsec:diagnosis}) and a taxonomy tree (Fig.~\ref{fig:taxonomy}) that links each symptom to the sections where it is discussed.}
\label{fig:case-studies-real}
\end{figure*}

\paragraph{\textbf{Analysis of Figure~\ref{fig:case-studies-real}.}}
Across tasks and evaluation settings, we consistently observe four dominant failure patterns that explain the majority of unsuccessful trials:
\begin{enumerate}[leftmargin=*]
    \item \textbf{Pre-grasp / Grasp errors:}  
    Examples in Figure~\ref{fig:case-studies-real}(a,b,c,e,h) predominantly reflect issues such as sub-centimeter XY/Z offsets and wrist-roll misalignment near small or thin features (e.g., sponge, zipper tab, waistband). These subtle misalignments often prevent stable contact or successful pickup.
    
    \item \textbf{Release / Placement timing errors:}  
    We also observe failures where the gripper releases too early or too late near receptacles, leading to failed placements or partial drops.
    
    \item \textbf{Instruction adherence failures:}  
    Figure~\ref{fig:case-studies-real}(d) shows a case where the command ``put sponge in pot'' triggers the \emph{Clean Dish} routine instead, reflecting dataset-prior–driven misalignment between language instructions and executed skills.
    
    \item \textbf{Trajectory / State drift:}  
    As shown in Figure~\ref{fig:case-studies-real}(f,g), failures also occur due to accumulated trajectory deviation over long horizons, particularly under visual occlusion or when the pulling direction diverges from the intended axis.
\end{enumerate}

In practice, multiple symptoms can co-occur, making causal attribution challenging. We emphasize the need for \emph{instrumented debugging} and \emph{model interpretability} to disentangle these intertwined factors.

Each failure type reflects a distinct underlying limitation in model behavior or sensory reliability. Pre-grasp and grasp errors primarily arise from visual grounding drift and depth sensing noise, which cause sub-centimeter offsets between the estimated and true object pose. Release and placement timing errors are often linked to controller latency and insufficient synchronization between perception and actuation, leading to premature or delayed gripper opening. Trajectory or state drift typically results from accumulated pose estimation errors and occlusion-induced visual uncertainty, where the model’s internal state diverges from the physical scene. These observations highlight that sensor precision, temporal coherence, and control bandwidth directly modulate the reliability of the learned policies across all models.

Across models, the primary failure modes align with our analyzed taxonomy. ACT mainly suffer from \textit{Trajectory / State drift} and unstable re-grasp attempts under OOD, reflecting its limited visual correction and error accumulation under occlusion. By contrast, $\pi_0$ maintains strong spatial stability but exhibits occasional \textit{depth-induced misplacements} (a \textit{Release / Placement} subtype) and \textit{Instruction adherence} lapses on sequential commands, attributed to depth perception noise and localization gaps. These patterns directly map to the model-level factors in our analysis: perception precision, release timing, state estimation, and language grounding. All observed failures are available at \url{<https://researchpaper253.github.io/sensys_vla_videos/>} for qualitative inspection.


\begin{figure}[t]
  \centering
  \resizebox{\linewidth}{!}{%
  \begin{forest} vla-tax-lr
    [VLA Failure Taxonomy
      [Task-level symptoms\\(see Sec.~\ref{subsec:task-failures})\label{fig:tax-task}
        [Pre-grasp / Grasp]
        [Release / Placement]
        [Trajectory / State drift]
        [Instruction adherence]
      ]
      [Model-level factors (abstract)\\(see Sec.~\ref{subsec:model-failures})\label{fig:tax-model}
        [{Perception misalignment}] 
        [Temporal inconsistency]
        [Language grounding gaps]
        [Policy generalization breakdown]
      ]
    ]
  \end{forest}}
  \caption{Revised failure taxonomy in the original left-to-right layout: root at left, two branches for task-level symptoms and model-level factors.}
  \label{fig:taxonomy}
\end{figure}
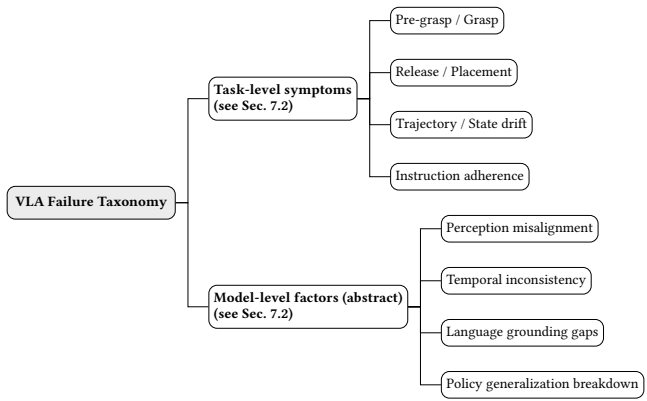

\section{Benchmark Results: Simulation}
\label{sec:results-sim}

To evaluate open-source checkpoints for OpenVLA--OFT and $\pi_0$, we designed a controlled environment in MuJoCo simulation. The environment includes a Franka Panda arm with a jaw gripper on a tabletop, interacting with three colored cubes: red, green, and blue. By default, the cubes are placed randomly on the table. We defined a set of tasks in this setting to probe specific language understanding skills:

\begin{itemize}
  \item \textbf{Color-Specific Pick \& Place.} Instruction: ``Pick up the red block.'' The robot must identify the correct cube by color and execute a grasp and lift. \emph{Evaluates}: basic instruction grounding and reliable pick execution.

  \item \textbf{Single-Step Stacking.} Instruction: ``Stack the red block on the green block.'' The robot picks up the first cube (red) and places it on the second (green). \emph{Evaluates}: reference resolution for two objects, understanding of the spatial relation ``on,'' and placement precision.

  \item \textbf{Two-Step Sequential Stacking.} Instruction: ``Stack the cubes in the order red, green, blue.'' The robot must perform ordered placements (e.g., stack red on green, then that pair on blue), maintaining the sequence specified in the language. \emph{Evaluates}: multi-step planning, short-horizon memory, and cumulative placement accuracy.
\end{itemize}

We evaluated the models in two conditions. In the In-Distribution (ID) condition, the scene is as described below. The cubes start in the default arrangement with standard lighting and no distractors. This setting is similar to what the model might have seen during fine-tuning. In the Out-of-Distribution (OOD) condition, we randomized the initial cube positions by shuffling their left--right order and varying their spacing. We also made slight adjustments to the lighting, adjusting the brightness and angle, and added a neutral-colored distractor object to the table. These perturbations test the robustness of the models' perception and language grounding to environmental changes.

A trial is considered successful based on the final outcome of the instruction. For a pick-up task, success means the specified cube is lifted at least 3cm off the table and held for at least 1second. For a stacking task, the first cube must be placed on top of the second and remain there for at least 2 seconds. The placed cube's base must be within 2.5cm horizontally of the top face of the lower cube, and the stack's height should be about two cube-heights (within $\pm$1\,cm). Each trial was given up to 10seconds to attempt the task. We focused on two models: the fine-tuned OpenVLA (using the optimized fine-tuning, denoted OpenVLA--OFT) and $\pi_0$. Our goal was to see how a foundation VLA model and our best-performing generalist compare in this pure language-directed setting. OpenVLA was evaluated zero-shot in this simulator, without additional fine-tuning on the specific simulated tasks, to assess its out-of-the-box generalization. Similarly, $\pi_0$ was not additionally fine-tuned on these instructions and was tested in a zero-shot manner, relying on its broad prior training.

\begin{table}[H]
    \centering
    \caption{Simulation SR (\%) for OpenVLA--OFT and $\pi_0$ under ID/OOD.}
    \label{tab:sim-openvla-pi0}
    \begin{tabular}{llcc}
        \toprule
        Task & Condition & OpenVLA--OFT & $\pi_0$ \\
        \midrule
        Pick up red cube       & ID  & 0\% & 60\% \\
                               & OOD & 0\% & 55\% \\
        Pick up green cube     & ID  & 0\% & 70\% \\
                               & OOD & 0\% & 68\% \\
        Pick up blue cube      & ID  & 0\% & 80\% \\
                               & OOD & 0\% & 75\% \\
        Stack red on green     & ID  & 0\% &  0\% \\
                               & OOD & 0\% &  0\% \\
        Stack blue on green    & ID  & 0\% &  0\% \\
                               & OOD & 0\% &  0\% \\
        \bottomrule
    \end{tabular}
\end{table}

\paragraph{\textbf{Results.}} Table~\ref{tab:sim-openvla-pi0} shows clear performance differences on the color-specified pick-and-place task. $\pi_0$ achieves a $60$--$80\%$ success rate in the in-distribution scenario and $55$--$75\%$ in the out-of-distribution scenario. In contrast, OpenVLA--OFT achieved $0\%$ success in both scenarios. 

Both models had a $0\%$ success rate on the single-step stacking task. $\pi_0$ often completed the grasp but failed in precise placement or release, causing the top block to fall; OpenVLA--OFT also achieved no successful stacking. The two-step sequential stacking task failed for both models and is not listed in the table.

\paragraph{\textbf{Analysis.}} These results show that $\pi_0$ demonstrates strong zero-shot language and generalization skills, transferring them to moderately perturbed settings. OpenVLA--OFT cannot perform any zero-shot tasks without further fine-tuning. The shared failure in stacking suggests that the main obstacle is not semantic understanding, but rather contact-intensive, dexterous manipulation: precise alignment and gentle release remain difficult without task-specific fine-tuning or specialized control mechanisms, such as higher-frequency closed-loop visual servoing or trajectory optimization. We also discovered that in multi-step stacking tasks, multi-step commands decreases the $\pi_0$'s first step SR. For example, in the single tasks, we use command ``pick up blue cube", the SR is around 75\%. However, when we use multi-step command ``put blue cube one top of red cube", the SR to pick up the blue cube drop to around 35\%. Overall, $\pi_0$ effectively followed instructions and generalized zero-shot on the pick-up task; both models need improvement for high-precision stacking. 

\begin{figure*}[t]  
  \centering
  \begin{subfigure}{0.25\linewidth}
    \includegraphics[width=\linewidth]{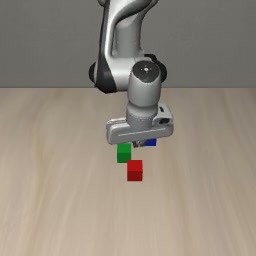}
    \caption{Pre-grasp---planar (XY) mis-grasp on color pick (OpenVLA--OFT)}
  \end{subfigure}\hfill
  \begin{subfigure}{0.25\linewidth}
    \includegraphics[width=\linewidth]{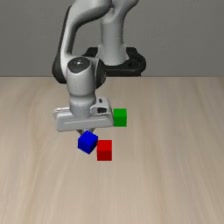}
    \caption{Pre-grasp---approach yaw/pitch error; push instead of pinch ($\pi_{0}$)}
  \end{subfigure}\hfill
  \begin{subfigure}{0.25\linewidth}
    \includegraphics[width=\linewidth]{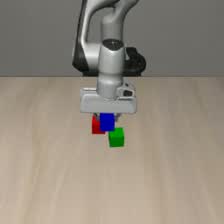}
    \caption{Placement---stack misalignment at release ($\pi_{0}$)}
  \end{subfigure}
  \caption{Simulation case studies aligned with our taxonomy: (a–b) pre-grasp geometry errors on color-specified picks; (c) placement misalignment during stacking. These failures arise from contact-level precision rather than instruction grounding.}
  \Description{Three simulated scenes: XY mis-grasp on a colored cube, orientation error causing a push, and a misaligned release during stacking.}
  \label{fig:case-studies-sim}
\end{figure*}

\subsection{Simulation Case Studies of Failure Modes}
\label{subsec:sim-case-studies}

\paragraph{\textbf{Analysis of Figure~\ref{fig:case-studies-sim}.}}
Figure~\ref{fig:case-studies-sim} demonstrates typical failure modes that observed in the simulator, which align closely with the taxonomy used in the real-world analysis. Unlike real-world scenarios where errors are caused by multiple factors such as depth noise, deformable objects, and lighting variations, the simulation environment highlights contact accuracy and controller timing as the primary bottlenecks, consistent with the aggregate results in Table~\ref{tab:sim-openvla-pi0}.

\begin{itemize}
    \item \textbf{(a) Pre-grasp planar (XY) mis-grasp on color pick (OpenVLA--OFT) — \emph{Pre-grasp/Grasp}.}
    Poor coarse spatial localization and limited fine alignment result in sub-centimeter XY offsets, causing the gripper close in empty space. This aligns with OpenVLA--OFT’s $0\%$ SR on all pick tasks without fine-tuning, indicating that geometric precision is the main failure factor. 
    \item \textbf{(b) Approach yaw/pitch error; push instead of pinch ($\pi_{0}$) — \emph{Pre-grasp/Grasp}.}
    $\pi_{0}$ reaches the correct target but comes with a small gripper orientation error; the fingers make tangential contact and push the object instead of a stable pinch. These observations partially explain why $\pi_{0}$ achieves correct semantic grounding but falls in final centimeter-level pose refinement
    \item \textbf{(c) Stack misalignment at release ($\pi_{0}$) — \emph{Release/Placement}.}
    In single-block stacking tasks, $\pi_{0}$ often succeeds at grasping but suffer from small angular misalignments upon release, causing top blocks to slide or topple.  
\end{itemize}

\textbf{Key Findings.} (i) The simulator reproduces the same task-level failure types observed in real experiments—where \textit{pre-grasp/grasp} and \textit{release/place} dominate—while clearly distinguishing them from command-following errors. (ii) $\pi_{0}$'s robust language-grounding capability explains its high grasp success rate; however, micro-pose alignment and release validation remain necessary for contact-intensive skills such as stacking. (iii) OpenVLA--OFT failures primarily occur during geometric alignment, suggesting that even moderate task-specific fine-tuning or closed-loop visual-servo control could yield significant performance gains. Collectively, these case studies further validate that improving centimeter-level pose correction and release-gating mechanisms represents the most direct pathway to bridging the gap between simulation and real-world performance for contact-sensitive tasks.


\section{Discussions with Failure Taxonomy, Diagnostic Analysis, and Remedy Insights}
\label{sec:failure-discussion}

This section provides a consolidated discussion on failure behaviors observed in both real-world and simulation experiments. We first describe how the failure taxonomy was constructed and used for analysis. We then present the detailed taxonomy itself, followed by in-depth diagnostic insights and potential remedies for each failure type. Finally, we discuss how these findings motivate tighter neuro-symbolic integration for interpretability and alignment.

\subsection{Constructing the Failure Taxonomy}
\label{subsec:taxonomy-construction}

We systematically reviewed all unsuccessful trials by replaying synchronized logs and multi-view videos. Each trial was assigned to one primary failure category based on its earliest causal error, as visualized in Figure~\ref{fig:taxonomy}.  
When multiple symptoms co-occurred (e.g., a grasp offset followed by trajectory drift), we annotated the first error leading to the eventual failure.  
This iterative process—refining categories through repeated cross-validation among annotators—produced a reproducible taxonomy that captures both mechanical and cognitive aspects of failure.

\subsection{Failure Taxonomy}
\label{subsec:taxonomy}

We group observed failures into two complementary levels: \textbf{task-level} (manifested at the behavior or skill execution level) and \textbf{model-level} (arising from perception, planning, or policy inference deficiencies). 

\subsubsection*{\textbf{Task-Level Failures}}
\phantomsection\label{subsec:task-failures}
\begin{itemize}[leftmargin=*]
    \item \textbf{Pre-grasp / Grasp errors.} The gripper approaches the object with an incorrect pose, leading to air grasps, slippage, or contact deflection. These are typically caused by sub-centimeter misalignments or wrist-roll errors when manipulating small features (e.g., sponge, zipper tab, waistband).
    \item \textbf{Release / Placement errors.} The gripper opens before the object is stably supported or misaligns with the target receptacle. The object consequently falls, tilts, or rests at an incorrect position (e.g., sponge landing on pot rim rather than inside).
    \item \textbf{Trajectory / State drift.} Accumulated perception or state-tracking errors cause deviations over long horizons, often after partial task success. This is common under visual occlusion or when interaction forces diverge from the intended trajectory.
    \item \textbf{Instruction adherence failures.} The behavior semantically mismatches the intended command—selecting the wrong object, relation, or order—despite syntactically valid motions.
\end{itemize}

\subsubsection*{\textbf{Model-Level Failures}}
\phantomsection\label{subsec:model-failures}
\begin{itemize}[leftmargin=*]
    \item \textbf{Perception misalignment.} Errors in visual grounding or depth estimation propagate to inaccurate spatial understanding.
    \item \textbf{Temporal inconsistency.} Control policies lose synchronization between motion planning and execution timing, resulting in instability during transitions.
    \item \textbf{Language grounding gaps.} The model misinterprets referring expressions or compositional relations in commands, reflecting a lack of shared semantics across modalities.
    \item \textbf{Policy generalization breakdown.} Model outputs collapse under unseen spatial or object configurations, indicating limited out-of-distribution adaptability.
\end{itemize}

\subsection{Failure Diagnosis}
\label{subsec:diagnosis}

To diagnose root causes, we analyze temporal event traces, gripper contact states, and camera views to identify the first divergence from ideal trajectories. This approach isolates causal failure points rather than post hoc symptoms.

\textbf{Pre-grasp/Grasp:} Root causes include coarse spatial priors, insufficient local alignment before closure, and lack of fine wrist-roll correction.  
\textbf{Release/Placement:} Timing inaccuracies and missing receptacle-frame checks lead to premature or misplaced release events.  
\textbf{Trajectory Drift:} Weak visual feedback loops and uncorrected state accumulation under occlusion cause compounding drift.  
\textbf{Instruction Adherence:} Limited multimodal grounding results in semantic mismatches between visual perception and language intention.

\begin{figure}[t]
    \centering
    \begin{tikzpicture}
        \begin{axis}[
            ybar,
            bar width=8pt,
            width=\linewidth,
            height=5cm,
            ymin=0, ymax=50,
            ylabel={\textbf{Failure Proportion (\%)}},
            symbolic x coords={
                Pre-grasp/Grasp,
                Release/Placement,
                Trajectory Drift,
                Instruction Adherence,
                Perception Misalignment,
                Temporal Inconsistency,
                Language Grounding Gap,
                Policy Breakdown
            },
            xtick=data,
            x tick label style={rotate=30, anchor=east, font=\footnotesize},
            ymajorgrids=true,
            grid style=dashed,
            nodes near coords,
            nodes near coords align={vertical},
            enlarge x limits=0.04
        ]
        \addplot coordinates {
            (Pre-grasp/Grasp, 32)
            (Release/Placement, 18)
            (Trajectory Drift, 15)
            (Instruction Adherence, 10)
            (Perception Misalignment, 9)
            (Temporal Inconsistency, 6)
            (Language Grounding Gap, 6)
            (Policy Breakdown, 4)
        };
        \end{axis}
    \end{tikzpicture}
    \vspace{-2mm}
    \caption{Distribution of failure categories across all evaluated trials (real and simulated). 
    Task-level failures dominate, with pre-grasp/grasp errors accounting for nearly one-third of all observed failures.}
    \label{fig:failure-distribution}
\end{figure}
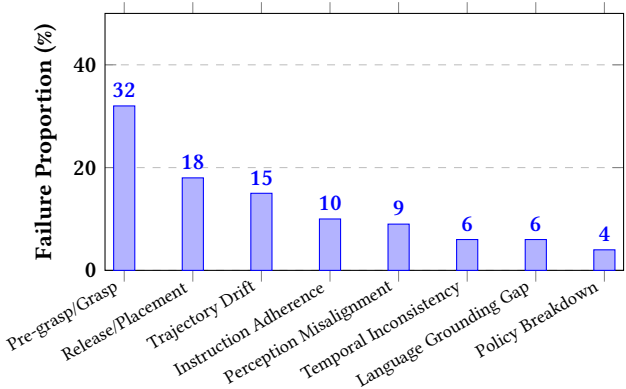

\noindent\textbf{Quantitative Summary.} 
Across $n{=}180$ annotated failure cases, task-level issues accounted for roughly \textbf{75\%} of failures, while model-level factors comprised the remaining \textbf{25\%}. 
Among task-level failures, \textbf{pre-grasp/grasp misalignments (32\%)} were the most frequent, followed by \textbf{release/placement timing errors (18\%)}, \textbf{trajectory drift under occlusion (15\%)}, and \textbf{instruction adherence mismatches (10\%)}. 
Within model-level failures, \textbf{perception misalignment (9\%)}, \textbf{temporal inconsistency (6\%)}, and \textbf{language grounding gaps (6\%)} were the main contributors. 
These statistics highlight that most failures stem from low-level geometric or temporal inaccuracies rather than high-level semantic misinterpretation.

\subsection{Remedy Insights and Discussion}
\label{subsec:remedy}

Based on our diagnostic analysis, several remedies are proposed to improve robustness and interpretability:

\begin{itemize}[leftmargin=*]
    \item \textbf{Pre-grasp alignment.} Introduce micro-alignment waypoints, wrist-roll normalization, and closed-loop force correction before grasp closure.
    \item \textbf{Release stability.} Enforce receptacle-frame contact validation (height/normal tests) and brief hold-for-stability delays before opening.
    \item \textbf{Trajectory correction.} Integrate subgoal resets, visual “clean-up” refinement near deformable regions, and chunked action outputs with partial overlap for long-horizon stability.
    \item \textbf{Semantic grounding.} Diversify linguistic expressions and spatial relations in demonstrations, incorporate contrastive negatives, and improve symbolic abstractions of perception.
\end{itemize}

\textbf{Broader Insight:}  
These observations reinforce the need for \emph{tight neuro-symbolic integration}—combining neural perception and symbolic reasoning to enable systematic interpretability and testability. Failures arising from latent misalignments could be better diagnosed if symbolic abstractions were embedded within neural control pipelines.

The recently proposed NeuroStrata framework~\cite{zheng2025neurostrata} and the Dual Neural-Symbolic Pipeline for UAV Landing~\cite{qian2025neurosymland} exemplify how structured symbolic reasoning can complement neural perception to enhance safety and interpretability. NeuroStrata~\cite{zheng2025neurostrata} proposes a hierarchical neurosymbolic framework for autonomous systems, ensuring deterministic and verifiable behavior by coupling the adaptability of neural networks with formal symbolic verification across perception and control layers. Similarly, NEUROSYMLAND~\cite{qian2025neurosymland} demonstrates the benefits of integrating probabilistic scene graphs and Scallop-based rule execution to achieve interpretable, verifiable, and mission-aware decision-making for UAVs in complex environments. 

These studies highlight a promising direction toward integrating symbolic structure into embodied intelligence. The proposed taxonomy provides a principled framework for future design by categorizing and diagnosing failures at multiple levels, it enables targeted improvements in model grounding, control robustness, and data collection strategies. These improvements can directly guide the next generation of benchmark protocols and and the optimization of training workflows for VLA models. 

\section{Conclusions}

This study systematically evaluated a task-specific imitation policy (ACT) and three VLA foundation models (OpenVLA, RDT-1B, and $\pi_{0}$) across real-world and simulation benchmarks under in-distribution and out-of-distribution conditions. 

Our results show that while ACT performs well in narrow, seen settings, it fails to generalize spatially or across objects. Large VLA models exhibit strong semantic grounding but poor contact precision, often missing grasps or releasing unstably. In contrast, $\pi_{0}$ achieves the best overall robustness, maintaining high success rates across all conditions. The number of demonstrations also has a strong effect, as doubling data substantially improves $\pi_{0}$'s success on long-horizon tasks.

Across all models, three dominant failure causes emerge—\emph{near-miss grasps, ill-timed releases, and trajectory drift}—highlighting the gap between high-level symbolic understanding and low-level motor execution. Tighter integration of neural perception and symbolic reasoning can bridge this gap: symbolic predicates can constrain and verify grasp and release conditions, while neural perception provides continuous feedback to refine spatial alignment and trajectory stability, jointly enabling safer, more interpretable, and more reliable task execution.

\bibliographystyle{ACM-Reference-Format}
\bibliography{references}

\balance
\end{document}